\pdfoutput=1

\documentclass[11pt]{article}

\usepackage{acl}

\usepackage{times}
\usepackage{latexsym}

\usepackage[T1]{fontenc}

\usepackage[utf8]{inputenc}

\usepackage{microtype}

\usepackage{inconsolata}

\usepackage{algorithm}
\usepackage{algorithmic}

\usepackage{graphicx}
\definecolor{lightgray}{rgb}{0.83, 0.83, 0.83}
\definecolor{rosegold}{rgb}{0.72, 0.43, 0.47}
\definecolor{ceruleanblue}{rgb}{0.16, 0.32, 0.75}
\definecolor{bittersweet}{rgb}{1.0, 0.44, 0.37}
\definecolor{amber(sae/ece)}{rgb}{1.0, 0.49, 0.0}
\definecolor{applegreen}{rgb}{0.55, 0.71, 0.0}
\definecolor{coralred}{rgb}{1.0, 0.25, 0.25}

\definecolor{gray_me}{rgb}{0.6, 0.6, 0.6}

\definecolor{capri}{rgb}{0.0, 0.75, 1.0}
\definecolor{amber}{rgb}{1.0, 0.75, 0.0}
\definecolor{blue_ppt}{rgb}{0.1914, 0.3359375, 0.9414}

\usepackage{enumitem}
\usepackage{array}
\usepackage{multirow}
\usepackage{pifont}
\newcommand{\cmark}{\ding{51}}%
\newcommand{\xmark}{\ding{55}}%
\usepackage{booktabs}
\usepackage{amsmath}
\usepackage{amssymb}
\usepackage{dashrule}
\usepackage{hyperref}

\title{Visually-Aware Context Modeling for News Image Captioning}

\author{Tingyu Qu\textsuperscript{1}, Tinne Tuytelaars\textsuperscript{2}, Marie-Francine Moens\textsuperscript{1} \\
  \textsuperscript{1}Department of Computer Science, KU Leuven \\
  \textsuperscript{2}Department of Electrical Engineering, KU Leuven \\
  \texttt{ \{tingyu.qu, tinne.tuytelaars, sien.moens\}@kuleuven.be}
   \\}

\begin{document}
\maketitle
\begin{abstract}

    News Image Captioning aims to create captions from news articles and images, emphasizing the connection between textual context and visual elements. 
    Recognizing the significance of human faces in news images and the face-name co-occurrence pattern in existing datasets, we propose a face-naming module for learning better name embeddings.
    Apart from names, which can be directly linked to an image area (faces), news image captions mostly contain context information that can only be found in the article.
    We design a retrieval strategy using CLIP to retrieve sentences that are semantically close to the image, mimicking human thought process of linking articles to images. 
    Furthermore, to tackle the problem of the imbalanced proportion of article context and image context in captions, we introduce a simple yet effective method Contrasting with Language Model backbone (CoLaM) to the training pipeline.
    We conduct extensive experiments to demonstrate the efficacy of our framework.
    We outperform the previous state-of-the-art (without external data) by 7.97/5.80 CIDEr scores on GoodNews/NYTimes800k.
    Our code is available at \href{https://github.com/tingyu215/VACNIC}{https://github.com/tingyu215/VACNIC}.
\end{abstract}

\section{Introduction}

Online news consumption heavily relies on news images as a key source of supplementary information alongside articles. 
These images, paired with engaging and informative captions, play a crucial role in capturing readers' attention. 
Typically, a news image illustrates a portion of the article, with the caption linking the image content to the article.
Ideally, readers should be able to grasp the essence of the news article by browsing through its images and their corresponding captions.

\begin{figure}[t]
    \centering
    \includegraphics[width=\columnwidth]{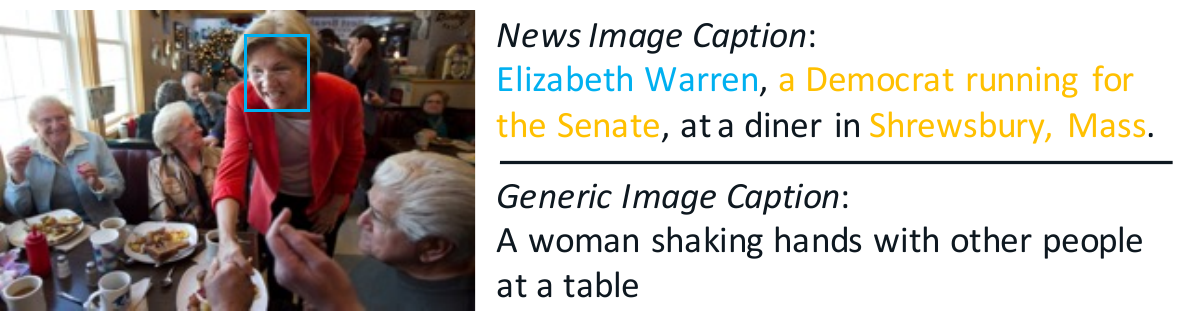}
    \caption{Two types of image captions. The image contains all context needed for the generic image caption, while in the news image caption, we find more named entities, including \textcolor{capri}{the name of a celebrity whose face appears in the image}, and \textcolor{amber}{context that is retrieved from the corresponding news article}. 
    Most of the context in the news image caption requires linking the image to the article.
    }
    \label{fig:example_image}
\end{figure}

News Image Captioning,
the task of generating a caption for an image using the contextual information derived from the corresponding article, 
contrasts with generic image captioning, where the image contains all necessary information for generating a descriptive sentence.
Figure \ref{fig:example_image} shows a generic image caption, and a news image caption from the GoodNews \cite{Biten_2019_CVPR} dataset.\footnote{The generic caption is generated with BLIP-2~\cite{li2022blip} More examples can be found Appendix~\ref{appendix:qualitative}.} 
In the news image caption, \textcolor{capri}{Elizabeth Warren} acts as a pivotal word. As a celebrity, \textcolor{capri}{Elizabeth Warren} can also be recognized from the image.
Furthermore, the news image caption contains \textcolor{amber}{context that is retrieved from the article}.
Moreover, all colored text in news image caption requires linking the image to the article, showing large imbalances in the proportion of article context and image context reflected in the caption.
In contrast, the generic image caption simply serves as a descriptive sentence of the image without any additional information.


Given the distinct nature of news image captions compared to generic image captions, an important question arises: How can visual inputs in News Image Captioning be used more effectively? 
Current methods primarily incorporate visual features from pretrained image encoders through cross-attention modules~\cite{Tran_2020_CVPR, yang-etal-2021-journalistic} or visual prefixes~\cite{zhang-et-al-mm22} to pre-trained language models.
This straightforward integration method is commonly applied in generic image captioning.
However, as also indicated by~\citet{zhang-wan-2023-exploring},there is a need for more effective utilization of images in News Image Captioning. 


We draw inspiration from studies on the human cognitive system, where studies indicate that faces uniquely capture human attention more than other objects in images~\cite{changing_faces}. 
Additionally, recognizing familiar faces enhances the recall of detailed "person knowledge," like personal traits and intentions~\cite{familiar_faces}. In News Image Captioning, this insight is particularly relevant, as news images frequently focus on human subjects. This understanding of how faces impact attention and memory guides our strategies for handling images centered around people.
In two commonly used News Image Captioning datasets, GoodNews~\cite{Biten_2019_CVPR} and NYTimes800k~\cite{Tran_2020_CVPR}, there is a notable pattern where over 56\% of samples feature both faces and names, while about 32\% have neither. All samples with significant faces in images also include names in their captions.\footnote{We provide more detailed statistics in Appendix~\ref{appendix:stat}.} This pattern, aligned with cognitive science's emphasis on the importance of faces in image perception, motivates the differentiation of faces from other objects in images for distinct treatment. 
We design a face-naming module to help the model to selectively attend to relevant names from the accompanying article. The face-naming module includes a prefix-augmented attention module~\cite{zhao-etal-2022-prefix-attn-domain} and is trained with a weakly supervised face-name alignment method~\cite{Qu_2023_WACV}.

Apart from the names, news image captions, unlike generic ones, often include contextual information (like "a Democrat running for the senate" in Figure~\ref{fig:example_image}) that cannot be directly linked to image areas. To generate these captions accurately, linking image content with relevant article segments is essential. We use a CLIP-based~\cite{pmlr-v139-radford21a} sentence retrieval strategy to find article sentences closely related to the image, aiding our caption generation process.



Moreover, news captions typically emphasize more context derived from the articles to engage readers and abstractly illustrate the article's content. 
Instead of explicitly modeling the image context and the article context, which requires detailed annotations, we propose Contrasting with Language Model backbone (CoLaM) which implicitly guide the model to prioritize article context. 
We align the embedding space of the multimodal models to the embedding space of their frozen language model backbones using a margin loss.
This approach ensures the model with multimodal inputs focus more effectively on article-related context.


To sum up, we introduce a novel framework for News Image Captioning that utilizes visual inputs differently than previous works. 
Our main contributions include:
\begin{enumerate}
    \item We are the first to introduce distinct modules tailored for different visual inputs in News Image Captioning, establishing the new state-of-the-art on two datasets.
    \item For visual inputs like faces that can be directly linked to textual context, 
    we design a face naming module 
    to utilize the commonly-occurred pattern of face-name co-occurrence in News Image Captioning datasets.
    For visual inputs that cannot be directly visually grounded,
    we design a sentence retrieval strategy using CLIP to bridge the gap between the article segments and the images. 
    The proposed modules result in significant improvement in performance.
    \item Addressing the imbalance between article and image context in the captions, we propose CoLaM, a universal method using a margin loss to enhance article context learning, further improving captioning performance.
\end{enumerate}


\begin{figure*}
\begin{center}
\includegraphics[width=\linewidth]{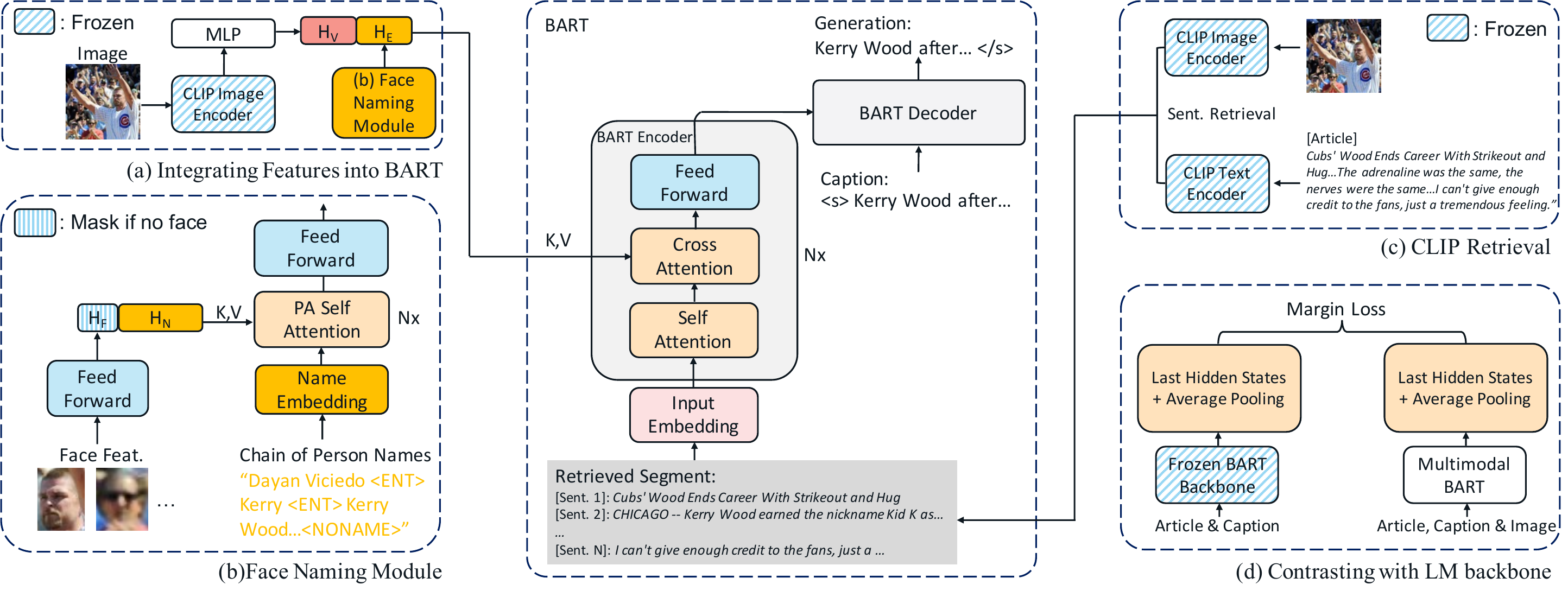}
\end{center}
   \caption{Method illustration. Our model is an encoder-decoder model built on BART (middle). 
   Our method consists of:
   (a) Integrating Features into BART: 
   In BART encoder, we concatenate visual ($H_V$) and name features ($H_E$) to obtain keys and values for the added cross-attention module;  
   (b) Face Naming Module: 
   We first get the embedding $H_N$ of the chain of person names in the article. 
   Then we prepend the face features $H_F$ to $H_N$ to obtain keys and values for 
   the prefix-augmented self-attention module;
   (c) CLIP Retrieval: We conduct sentence retrieval using CLIP to learn from more accurate article context.
   (d) Contrasting with LM backbone (CoLaM): We contrast the multimodal BART with frozen pure-text BART to force the model to focus more on the article context.
   }
\label{fig:model}
\end{figure*}

\section{Related Work}

In contrast to generic image captioning methods \cite{Karpathy_2015_CVPR, Donahue_2015_CVPR, Vinyals_2015_CVPR, pmlr-v37-xuc15, Anderson_2018_CVPR, Lu_2017_CVPR}, which rely solely on images as input, News Image Captioning takes both images and news articles as input. 
It dictates that models should prioritize captions that not only depict image content, but also summarise the corresponding article segments.

Early approaches to the task focus on learning the representation of a news article and its connection with an image, including utilizing n-gram language models for extracting phrases seen in the article \cite{TPAMI13-caption}, or building an encoder-decoder based architecture with VGG \cite{iclr15-vgg} for encoding the image, Word2Vec \cite{NIPS2013-word2vec} for encoding the article, and an LSTM as caption decoder \cite{TPAMI-breakingnews}. They all fail to achieve satisfactory performance. \citet{Biten_2019_CVPR} propose the first large-scale dataset GoodNews for the task, and a two-stage template based captioning method.
Following \citet{Biten_2019_CVPR}, several works adopt transformer-based models with different types of features like Places 365~\cite{TPAMI-places} used by \citet{yang-okazaki-2020-image} or face and object features used by the Tell model~\cite{Tran_2020_CVPR}.


Since then, the focus of the community has shifted to learning better entity representations. VisualNews~\cite{liu-etal-2021-visual} adopts a multi-head attention-on-attention module~\cite{Huang_2019_ICCV} and visual selective gates. 
JoGANIC~\cite{yang-etal-2021-journalistic} brings external knowledge from a Wikipedia database to train an entity embedding. 
On top of Tell, \citet{zhou-etal-2022-focus} show that an entity-aware retrieval method can improve the performance further. \citet{zhang-et-al-mm22} propose a prompt-based model NewsMEP with pre-trained BART \cite{lewis-etal-2020-bart} and CLIP features as the backbone.
NewsMEP follows ClipCap \cite{mokady2021clipcap} to generate visual prompts from CLIP representations. 
Instead of differentiating between different types of entities like our method, all entities are treated equally by NewsMEP, where a bi-LSTM is trained to learn the most important entities from the article as part of prompts to the decoder of NewsMEP.
We design different modules for different types of textual context information based on their connection to the various types of visual inputs, which yields superior performance than NewsMEP. We present a detailed comparison in Section~\ref{sec:results}.


Finally, in contrast to our work, where no additional paired datasets are used, recent works~\cite{ZHANG2022108429, rajakumar-kalarani-etal-2023-lets} explore the use of extra large-scale datasets on the task and obtain satisfactory results.


\section{Methodology}

\subsection{Model Architecture}

We present our model in Figure \ref{fig:model},
which is an encoder-decoder model built upon the generative pre-trained language model BART \cite{lewis-etal-2020-bart}. 
We add a cross-attention module to the BART encoder to integrate the visual and name features,
while keeping the BART decoder unchanged. 
To obtain the name features, we use a prefix-augmented self-attention module \cite{zhao-etal-2022-prefix-attn-domain} to softly select person names in the article that are similar to the detected faces. 
We also design a retrieval strategy using CLIP to retrieve article sentences that capture crucial context that cannot be directly inferred from the images. The retrieved segments serve as input to our model.
Section~\ref{sec:learning} presents our training pipeline.

\subsection{Integrating Features into BART}

We add a cross-attention module to the BART encoder to incorporate visual ($H_V$) and name features ($H_E$).
We use a simple MLP network as suggested by \citet{mokady2021clipcap} to get visual representations denoted as $H_V$ from the frozen CLIP image encoder.
As shown in Figure \ref{fig:model} (a), 
We concatenate $H_V$ and $H_E$ as $[H_V;H_E]$, which is linearly transformed to get the keys $K_A$ and values $V_A$. Together with the linearly transformed query $Q_A$ from the article hidden states, in each encoder layer, we compute the cross-attention as $\text{softmax}(Q_A {K_A}^{T}/\sqrt{d_H})V_A$, with $1/\sqrt{d_H}$ as the scaling factor.

\subsection{Face Naming Module}

As stated before, there is a strong face-name co-occurrence pattern in the News Image Captioning datasets.
We design a novel face naming module to learn a face-aware representation of person names,
as shown in Figure~\ref{fig:model} (b). 
Given a chain of person names in the article (e.g. "Dayan Viciedo $\langle$ENT$\rangle$ Kerry $\langle$ENT$\rangle$ ..." with $\langle$ENT$\rangle$ being a special token used as separator), we first compute the embedding $H_N$.
We generate face embedding $H_F$ by passing the face features\footnote{Face features are provided in the datasets.} through a feed-forward layer.
Then we prepend $H_F$ to $H_N$ as $[H_F; H_N]$, which is linearly transformed to keys $K_N$ and values $V_N$.
Together with the linearly transformed queries $Q_N$ from $H_N$, we compute the prefix-augmented self attention (PA self attention) as $\text{softmax}(Q_N {K_N}^{T}/\sqrt{d_H})V_N$, with $1/\sqrt{d_H}$ as the scaling factor. Finally we use a feed-forward layer to generate the name features $H_E$ of fixed length from the attended name embeddings.

With $H_F$, we can control the utilization of the contextual information from the faces. 
For images with no faces,
we mask $H_F$, resulting in a conventional self attention. If faces occur, the chain of person names receives contextual information from the faces through PA self attention. We detail the learning of this module in the Section~\ref{sec:learning}.




\subsection{CLIP Retrieval}
Unlike for the names and faces, where a clear connection between text and image can be found, it is difficult to find a direct connection between text and image for context that cannot be directly visually grounded. 
For such context, readers tend to use image contents to retrieve relevant information from the article.
In an effort to simulate this cognitive process, we use CLIP to retrieve sentences that are semantically closest to the image representation generated by the CLIP image encoder (measured by cosine similarity). To make sure we include enough global information, we also add the first three sentences of each article segment if they are not part of the retrieved sentences, and keep the original sentence ordering.

\subsection{Learning}
\label{sec:learning}

In this section we detail the learning process of our model.
Our full model is trained with a face naming loss, a margin loss (CoLaM) and a caption generation loss.




\noindent \textbf{Face naming loss}
Inspired by \citet{Qu_2023_WACV}, we adopt a symmetric contrastive loss to align faces in images to names in captions during training. 
Given $m-1$ names in a caption, we add an additional $\langle$NONAME$\rangle$ token. We denote the name embedding of each name as 
$H_{N,gt}^j$, $j = 1, 2, \ldots, m$. 
Since we only use the ground truth names during training for learning better face representations, 
we apply stop gradient to the name embedding layer while computing the loss. We denote the name embeddings with stop gradient as 
$\Tilde{H}_{N,gt}^j$, $j = 1, 2, \ldots, m$.
For the corresponding image with $n$ faces, we extract hidden states of faces $H_F$ from the last layer of the face naming module in our model. 
We denote the representation of each face as $H^i_F$ for $i = 1, 2, \ldots, n$.
For face set $F$ and name set $N$, 
we adopt the face-to-name contrastive loss as:

\begin{equation}
    \mathcal{L}_{f,n} = -\log \frac{e^{sim_d(F,N)}}{\sum_{F_k \in batch} e^{sim_d(F_k ,N)} }
    \label{eq:face_name}
\end{equation}

where $sim_d(F, N) = \frac{1}{n} \sum_{i=1}^{n} \max_{j} A_{i,j}$, with $A_{i,j} =  (H^i_F)^T \cdot \Tilde{H}_{N,gt}^j$, for $i=1,2,\ldots, n$, $j=1,2,\ldots, m$.

Similarly, we obtain the name-to-face contrastive loss as:

\begin{equation}
    \mathcal{L}_{n,f} = -\log \frac{e^{sim_d(N,F)}}{\sum_{N_k \in batch} e^{sim_d(N_k ,F)} }
    \label{eq:name_face}
\end{equation}

where $sim_d(N, F) = \frac{1}{m} \sum_{i=1}^{m} \max_{i} A_{j,i}$, with $A_{j,i} = (\Tilde{H}_{N,gt}^j)^T \cdot H^i_F $ , for $j=1,2,\ldots, m$, $i=1,2,\ldots, n$.

Combining Equation~\ref{eq:face_name} and \ref{eq:name_face}, we obtain a symmetric face naming loss as:

\begin{equation}
    \mathcal{L}_{f\leftrightarrow n} = \mathcal{L}_{f,n} + \mathcal{L}_{n,f}
\end{equation}

\noindent \textbf{CoLaM} 
The key idea of our Contrasting with Language Model (LM) backbone (CoLaM) is to guide the multimodal LMs to learn to focus more on the context
from the news articles through a margin loss by utilizing the LM backbones.

Figure~\ref{fig:model} (d) shows the simplified modeling process of our CoLaM. 
Specifically,
let $h_{lm}$ be a generative LM  (e.g. BART~\cite{lewis-etal-2020-bart}) backbone, and $h_{mm}$ be the generative multimodal LM built upon it.
We extract the last hidden states $C_{lm}$ and $C_{mm}$ for the generated text from the decoders of $h_{lm}$ and $h_{mm}$, respectively.
We compute the margin loss as:

\begin{equation}
\resizebox{.89\hsize}{!}{
    $\mathcal{L}_{m} = \frac{1}{B} \sum_{i} \max\{0, \Delta- cos(pool(C_{lm}^i), pool(C_{mm}^i) )\}$
}
\end{equation}

where $B$ is the batch size, $\Delta$ is the margin hyperparameter, $cos(\cdot)$ denotes the cosine similarity, $pool(\cdot)$ is the average pooling operation which takes into account the masking.

By applying average pooling, we obtain the global representations from $C_{lm}$ and $C_{mm}$, which is used to measure the cosine similarity between the two representations. As we freeze the text-only LM backbone, optimizing $\mathcal{L}_{m}$ is equivalent to adding a constraint to the multimodal LM. This constraint ensures the multimodal LM to put more emphasis on the news articles.
As shown before, a news image caption often contains more context from the article than from the image. Our CoLaM is a universal method for improving context modeling abilities of existing models, and can be seamlessly integrated into existing models. We further discuss the use of CoLaM in Section~\ref{sec:results} and Appendix~\ref{appendix:colam}.

\noindent \textbf{Caption generation loss} Given a news image and article pair, we minimize the negative log likelihood for caption generation as:

\begin{equation}
    \mathcal{L}_{cap} = -\sum_{t=1}^T \log p(y_t | y_{<t};\theta )
    \label{eq:lm_loss}
\end{equation}

where $y_t$ denotes target caption token at time step $t$, $y_{<t}$ denotes the current token sequence and $\theta$ represents the learned parameters of the model.

Finally, we train our model with the loss as:

\begin{equation}
    \mathcal{L} = \mathcal{L}_{cap} + \mathcal{L}_{f\leftrightarrow n} + \alpha \mathcal{L}_m
    \label{eq:total_loss}
\end{equation}
where $\alpha$ is the hyperparameter.

\section{Experiments}

\begin{table*}
\centering
\resizebox{0.9\linewidth}{!}{
\begin{tabular}{clccccccccc}
\toprule
& \multirow{2}{*}{Method} &  \multirow{2}{*}{Extra Data\textsuperscript{*}} & \multicolumn{4}{c}{Caption Generation\textsuperscript{\dag}} & \multicolumn{2}{c}{Named Entities\textsuperscript{\dag}} \\

&                         &         & B &  M & R & C & P & R  \\ \hline
\multirow{10}{*}{\rotatebox{90}{GoodNews}} 
& Avg+CtxIns \cite{Biten_2019_CVPR}  & \xmark & 0.89 & 4.37 & 12.20 & 13.10 & 8.23 & 6.06 \\ 
& Tell \cite{Tran_2020_CVPR}  & \xmark  & 6.05 & 10.30 & 21.40 & 53.80 & 22.20 & 18.70 \\
& VisualNews \cite{liu-etal-2021-visual}  & \xmark  & 6.10 & 8.30 & 21.60 & 55.40 & 22.90 & 19.30 \\

& \color{gray_me} JoGANIC \cite{yang-etal-2021-journalistic} & \color{gray_me} \cmark  & \color{gray_me} 6.83 & \color{gray_me} 11.25 & \color{gray_me} 23.05 & \color{gray_me} 61.22 & \color{gray_me} 26.87 & \color{gray_me} 22.05 \\

& \color{gray_me}Tell + Focus! \cite{zhou-etal-2022-focus} & \color{gray_me} $\bullet$ & \color{gray_me} 6.30 & \color{gray_me} $\diagup$ & \color{gray_me} 23.00 & \color{gray_me} 60.30 & \color{gray_me} 24.20 & \color{gray_me} 20.90 \\

& \color{gray_me} DiscExt CapGen \cite{ZHANG2022108429} & \color{gray_me} \cmark & \color{gray_me} 7.94 & \color{gray_me} 13.97 & \color{gray_me} 28.68 & \color{gray_me} 64.51 & \color{gray_me} 29.69  & \color{gray_me} 27.37 \\ 

& \citet{rajakumar-kalarani-etal-2023-lets} & \color{gray_me} \cmark & \color{gray_me} 7.14 & \color{gray_me} 11.21 & \color{gray_me} 24.30 & \color{gray_me} 72.33 & \color{gray_me} 24.37 & \color{gray_me} 20.09 \\

& NewsMEP \cite{zhang-et-al-mm22} &  \xmark  & \underline{8.30} & \underline{12.23} & \underline{23.17} & 63.99 & 23.43  & \underline{23.24} \\ \cline{2-9}

& $\text{Ours}_{\text{base}}$ (w/ $\text{BART}_{\text{base}}$) & \xmark & 7.20 & 11.00 & 21.97 & \underline{65.42} & \underline{24.15} & 22.18\\ 

& $\text{Ours}_{\text{large}}$, (w/ $\text{BART}_{\text{large}}$) & \xmark & 
\textbf{8.60} & \textbf{12.39} & \textbf{23.38} & \textbf{71.96} & \textbf{24.30} & \textbf{25.54} \\


\midrule
\multirow{10}{*}{\rotatebox{90}{NYTimes800k}} 
& Tell \cite{Tran_2020_CVPR} & \xmark  & 6.30 & 10.30 & 21.70 & 54.40 & 24.60 & 22.20 \\

& VisualNews \cite{liu-etal-2021-visual} & \xmark  & 6.40 & 8.10 & 21.90 & 56.10 & 24.80 & 22.30 \\

& \color{gray_me} JoGANIC \cite{yang-etal-2021-journalistic} & \color{gray_me} \cmark  & \color{gray_me} 6.79 & \color{gray_me} 10.93 & \color{gray_me} 22.80 & \color{gray_me} 59.42 & \color{gray_me} 28.63 & \color{gray_me} 24.49 \\

& \color{gray_me} Tell + Focus!(only CLIP) \cite{zhou-etal-2022-focus} & \color{gray_me} $\bullet$ & \color{gray_me} 6.40 & \color{gray_me} $\diagup$ & \color{gray_me} $\diagup$ & \color{gray_me} 57.50 & \color{gray_me} 25.70 & \color{gray_me} 22.70 \\

& \color{gray_me} Tell + Focus! \cite{zhou-etal-2022-focus} & \color{gray_me} $\bullet$ & \color{gray_me} 7.00 & \color{gray_me} $\diagup$ & \color{gray_me} 22.90 & \color{gray_me} 63.60 & \color{gray_me} 29.80 & \color{gray_me} 25.90 \\ 

& \color{gray_me} DiscExt CapGen \cite{ZHANG2022108429} & \color{gray_me} \cmark  & \color{gray_me} 7.57 & \color{gray} 12.64 & \color{gray_me} 25.67 & \color{gray_me} 62.31 & \color{gray_me} 30.04  & \color{gray_me} 25.53 \\ 

& \citet{rajakumar-kalarani-etal-2023-lets} & \color{gray_me} \cmark & \color{gray_me} 7.54 & \color{gray_me} 11.27 & \color{gray_me} 23.28 & \color{gray_me} 66.41 & \color{gray_me} 28.21 & \color{gray_me} 23.25 \\

& NewsMEP~\cite{zhang-et-al-mm22} & \xmark  & \textbf{9.57} & \textbf{13.02} & \textbf{23.62} & \underline{65.85} & 26.61  & \underline{28.57} \\ \cline{2-9} 

& $\text{Ours}_{\text{base}}$ (w/ $\text{BART}_{\text{base}}$) & \xmark & 7.87 & 11.19 & 21.95 & 64.64 & \textbf{26.98} & 25.33 \\ 

& $\text{Ours}_{\text{large}}$ (w/ $\text{BART}_{\text{large}}$) & \xmark &  
\underline{9.24} & \underline{12.57} & \underline{23.44} & \textbf{71.65} & \underline{26.88} & \textbf{28.59} \\

\bottomrule
\end{tabular}
}
\caption{Performance comparison with state-of-the-art methods. We highlight the best scores and underline the second best scores of models that do not use (a) extra data, or (b) additional pre-trained models other than the language model - vision backbones. 
\textsuperscript{*}: The use of extra data can be found in Appendix~\ref{appendix:implementation}.
\textsuperscript{\dag}: B: BLEU-4; R: ROUGE-L; M: METEOR; C: CIDEr; P: Precision; R: Recall. We adopt the same abbreviation in all tables.}
\label{tab:main_results}
\end{table*}



\subsection{Implementation Details}

We conduct experiments on two large-scale News Image Captioning datasets, namely GoodNews\cite{Biten_2019_CVPR} and NYTimes800k~\cite{Tran_2020_CVPR}. The details of the datasets are presented in Appendix~\ref{appendix:stat}.
Following the same experimental settings as previous works \cite{Tran_2020_CVPR, yang-etal-2021-journalistic},
we train our full model for 16/9 epochs on GoodNews/NYTimes800k.
We set the batch size to 32, the learning rate to 1e-5, and warm up for the first 5\% steps.
We adopt the AdamW optimizer \cite{loshchilov2018decoupled} with $\beta_1 = 0.9$, $\beta_2 = 0.999$ and $\epsilon=1e-8$, and apply weight decay of 0.01 to all weights as regularization. We clip the gradient norm at 0.1.
Following \citet{zhang-et-al-mm22}, a frozen CLIP-ViT-B/16 is used as image encoder.
The MLP network in visual feature generation module consists of two linear layers with hyperbolic tangent activation in between.
The same embedding layer structure as in BART is adopted for the name embedding.
We add two special tokens $\langle$ENT$\rangle$ and $\langle$NONAME$\rangle$, for separating names in a chain of person names, and acting as a "NONAME" token as suggested by \citet{Qu_2023_WACV}, respectively.
We set the length of the visual features and name features to be 20.
We detail the implementation to obtain such features in Appendix~\ref{appendix:implementation}.
During training, we restrict the number of tokens in articles and captions to be 512 and 100, respectively. We set $\alpha=2.0$ and $\Delta=1.0$ for CoLaM.
During inference, we use beam search with beam size of 5.
We provide more implementation details in Appendix~\ref{appendix:implementation}.

\subsection{Evaluation Metrics}

We follow the same evaluation pipeline as in previous works \cite{Biten_2019_CVPR, Tran_2020_CVPR, yang-etal-2021-journalistic, zhao-etal-2021-entity-graph, zhang-et-al-mm22}. To measure the overall quality of generated captions, we use BLEU-4 \cite{bleu-acl02}, ROUGE-L \cite{lin-2004-rouge}, METEOR \cite{denkowski-lavie-2014-meteor} and CIDEr scores \cite{Vedantam_2015_CVPR}. For BLEU-4 and ROUGE-L, every word contributes to the metric equally. METEOR focuses on synonym matching and lemmatization, which are seldomly found for named entities. CIDEr uses TF-IDF weighting to put more emphasis on rare words, e.g. named entities \cite{Biten_2019_CVPR, kilickaya-etal-2017-evaluating, elliott-keller-2014-comparing}. So following previous works, we also consider CIDEr as the most suitable one for the task.
We also use precision and recall to evaluate the quality of generated named entities.

\subsection{Results}
\label{sec:results}

We report the overall performance and the main ablation studies in this part. 
We present more results on human evaluation, text summarization, additional ablation studies, large vision-language models, in-depth study of CoLaM and qualitative analyses in Appendix~\ref{appendix:human}, \ref{appendix:text_summ}, \ref{appendix:more_ablation}, \ref{appendix:large_vlms}, \ref{appendix:colam} and \ref{appendix:qualitative}, respectively.

\noindent \textbf{Overall Performance}

We present the overall performance of our model in Table \ref{tab:main_results}. 
With a much smaller language model (LM) $\text{BART}_{\text{base}}$,\footnote{NewsMEP: $\text{BART}_{\text{large}}$; Tell\& JoGANIC: $\text{RoBERTa}_{\text{large}}$} we already achieve a competitive performance on both GoodNews and NYTimes800k datasets, when compared to the previous state-of-the-art (SOTA) model NewsMEP, which uses $\text{BART}_{\text{large}}$ as backbone.
When we increase the LM size to $\text{BART}_{\text{large}}$, we establish a new SOTA in terms of CIDEr scores and outperform NewsMEP by a large margin  (+6 points).
Our model also yields new SOTA entity scores on both datasets leading to more trustworthy captions.
Compared to our model, NewsMEP is constructed with the same vision backbone (CLIP-ViT-B/16) and LM ($\text{BART}_{\text{large}}$).
NewsMEP also adopts a prefix-augmented attention module to integrate visual and entity information into the model. 
It learns to select entities through the interaction between image and article representations from the encoder in $\text{BART}_{\text{large}}$.
However, NewsMEP fails to utilize the characteristics of different types of visual inputs, which should be treated differently as in our framework. 
By considering 
all visual inputs equally, NewsMEP lacks in linking rare words in articles or captions to visual inputs, resulting in much lower CIDEr scores when compared to our method.
Unlike generic image captioning, where the goal is to make a simple descriptive caption to the image, News Image Captioning requires the generated captions to capture the essence of both images and articles. In that case, CIDEr as a evaluation metric which put more emphasis on rare words should be prioritized. 
Our method obtains the highest CIDEr scores, showing its efficacy.

Focus!~\cite{zhou-etal-2022-focus} is a sentence retrieval method. 
Combining Focus! with Tell, relatively high CIDEr scores can be attained. 
Although no extra data sources are used, Focus! uses CLIP and OpenNRE~\cite{han-etal-2019-opennre} (a pre-trained domain specific relation extraction model) to perform sentence retrieval, on top of the LM\&vision backbones from Tell. 
Without OpenNRE, the CIDEr score of Tell + Focus! on NYTimes800k drops from 63.60 to 57.50, indicating the biggest gain of their method comes from the use of OpenNRE.
However,
we are more interested in the question: Without additional domain specific pre-trained models, how can we explore the connections between images, articles and captions of the given dataset?
Experimental results show the merit of our method, 
which is also demonstrated in the ablation study of the different components of our model (see below).

There are also three methods, namely JoGANIC \cite{yang-etal-2021-journalistic} DiscExt CapGen \cite{ZHANG2022108429} and \citet{rajakumar-kalarani-etal-2023-lets}, that use extra data sources in their framework. 
Without using extra data, our method significantly achieves higher CIDEr scores as compared to the former two methods, and yields comparable or better performance than \citet{rajakumar-kalarani-etal-2023-lets}, which uses the extra News Image Captioning dataset VisualNews~\cite{liu-etal-2021-visual} containing more than 1.2 million samples.
Our method generates better captions by exploring the News Image Captioning datasets in a better way.

\noindent \textbf{Ablation Study on Model Components}

We present results for the ablation study on the different components of our model in Table~\ref{tab:ablation_model}.\footnote{Limited to resources, we conduct the ablation studies using $\text{BART}_{\text{base}}$ as the backbone LM.}

\begin{table}[!htbp]
    \centering
    \resizebox{\columnwidth}{!}{
    \begin{tabular}{cccccccccccc}
    \toprule
        &  \multirow{2}{*}{Model} &  \multirow{2}{*}{VF} &  \multirow{2}{*}{NF} &  \multirow{2}{*}{RS} &\multirow{2}{*}{CoLaM} & \multicolumn{4}{c}{Caption Generation} & \multicolumn{2}{c}{Named Entities} \\
        & & & & & & B & M & R & C & P & R \\
    \hline
        \multirow{6}{*}{\rotatebox{90}{\small GoodNews}}  & $\langle$1$\rangle$ & & & & & 6.14 & 9.69 & 19.51 & 55.24 & 21.17 & 18.81 \\
         & $\langle$2$\rangle$ & \cmark & & & & 6.59 & 10.17 & 20.33 & 58.55 & 22.20 & 20.46 \\
         & $\langle$3$\rangle$ & \cmark & \cmark & &  & 6.81 & 10.54 & 21.17 & 61.73 & 22.86 & 21.02 \\
         & $\langle$4$\rangle$ & \cmark &  & \cmark &  & 6.81 & 10.42 & 20.88 & 60.41 & 22.96 & 20.76 \\
         & $\langle$5$\rangle$ & \cmark & \cmark & \cmark & & 7.00 & 10.75 & 21.79 & 64.07 & 23.69 & 21.50 \\
         & $\langle$6$\rangle$ & \cmark & \cmark & \cmark & \cmark & 
          \textbf{7.20} & \textbf{11.00} & \textbf{21.97} & \textbf{65.42} & \textbf{24.15} & \textbf{22.18} \\
    \midrule
        \multirow{6}{*}{\rotatebox{90}{\scriptsize NYTimes800k}}  & $\langle$1$\rangle$ & & & & & 6.63 & 9.89 & 19.14 & 51.76 & 22.30 & 21.63 \\
         & $\langle$2$\rangle$ & \cmark & & & &  6.75 & 10.12 & 19.74 & 54.45 & 24.04 & 22.47 \\
         & $\langle$3$\rangle$ & \cmark & \cmark & & & 7.18 & 10.63 & 20.81 & 59.07 & 25.70 & 23.78 \\
         & $\langle$4$\rangle$ & \cmark &  & \cmark & & 7.31 & 10.63 & 20.97 & 61.52 & 26.01 & 23.59 \\
         & $\langle$5$\rangle$ & \cmark & \cmark & \cmark & & 7.53 & 10.98 & 21.63 & 63.95 & 26.94 & 24.72 \\
         & $\langle$6$\rangle$ & \cmark & \cmark & \cmark & \cmark & 
         \textbf{7.87} & \textbf{11.19} & \textbf{21.95} & \textbf{64.64} & \textbf{26.98} & \textbf{25.33} \\

    \bottomrule
    \end{tabular}
    }
    \caption{Effects of different components of our model on qualities of generated captions. Model $\langle$1$\rangle$: $\text{BART}_{\text{base}}$; VF: visual features; NF: name features from face naming; RS: retrieved segments. Model $\langle$5$\rangle$: $\text{Ours}_{\text{base}}$.}
    \label{tab:ablation_model}
\end{table}

\noindent \textbf{Visual Features} When we discard image inputs, the task becomes a purely textual sequence-to-sequence 
problem. 
$\text{BART}_{\text{base}}$ (Model $\langle$1$\rangle$) can achieve fairly good results in this scenario on two datasets. However, it always generates the same caption for different images of an article.
The addition of the visual features in Model $\langle$2$\rangle$ mitigates the problem.
We observe consistent improvements in all evaluation metrics as shown in Table \ref{tab:ablation_model}. 

\noindent \textbf{Face Naming Module} On top of 
Model $\langle$2$\rangle$,
when we add the name features learned from our face naming module (Model $\langle$3$\rangle$), we observe significant improvement on all the evaluation metrics from both datasets, especially regarding the CIDEr score (58.55$\rightarrow$61.73 on GoodNews, 54.45$\rightarrow$59.07 on NYTimes800k). When both the visual features and name features are added to $\text{BART}_{\text{base}}$, we already achieve CIDEr scores higher than some models based on $\text{RoBERTa}_{\text{large}}$ (e.g., Tell with 53.80/54.40 CIDEr on GoodNews/NYTimes800k), or comparably to models that use extra external data (e.g., JoGANIC with 61.22/59.42 CIDEr on GoodNews/NYTimes800k). 

\noindent \textbf{CLIP Retrieval} We further improve the quality of the generated captions by retrieving sentences from the articles. In this way the model learns to focus on different segments for captioning different images. As shown in Table \ref{tab:ablation_model}, we improve the CIDEr scores from 61.73 to 64.07 on GoodNews, 59.07 to 63.95 on NYTimes800k. Apart from the improvements in caption generation evaluation metrics, we also improve the precision of all entity names generated in the captions 
on two datasets
after adding the retrieval component into our method.

We also observe models with VF+NF (Model $\langle$3$\rangle$) and VF+RS (Model $\langle$4$\rangle$) reach comparable performance, both significantly surpassing the model with only VF.
However, they still fall short of the model combining VF+NF+RS (Model $\langle$5$\rangle$).

\noindent \textbf{CoLaM} Finally, with the addition of CoLaM, the base version of our full model (Model $\langle$6$\rangle$) further improvements the performance of Model $\langle$5$\rangle$ on all metrics. 
It shows that the imbalanced proportion of context from articles and images in the captions can be a big problem for News Image Captioning models.
We present more in-depth analyses of the behavior of CoLaM in Appendix~\ref{appendix:colam}, together with the results of CoLaM with other model architectures to show its potential of being the universal add-on for News Image Captioning models.

\noindent \textbf{Ablation Study for Entity Generation}

The different components of our model also affect the generation of different types of entity names. We present the precision and recall scores of the three most commonly occurring entity types\footnote{PERSON: people; GPE: countries, cites, states; ORG: companies, agencies, etc.} in Table \ref{tab:ablation_ent_type}.
We have designed the face naming module to force the model to focus on the correct PERSON-type entities (names) which can be visually grounded from the images. 
As shown in Table \ref{tab:ablation_ent_type}, by adding the name features, we observe a significant improvement in both precision (28.00 $\rightarrow$ 29.22 on GoodNews, 32.86 $\rightarrow$ 37.11 on NYTimes800k) and recall (24.13 $\rightarrow$ 25.91 on GoodNews, 29.41 $\rightarrow$ 33.17 on NYTimes800k) of PERSON-type entity names, which shows the effectiveness of our face naming module. Interestingly, we also observe improvements in entity scores for other types of entities after adding the name features, for instance the recall of GPE (27.69 $\rightarrow$ 28.34 on NYTimes800k). We think this is due to the large improvement in predicting PERSON-type entity names, which leads to more accurate context modeling in the articles, and to generating captions of higher quality. 
We also observe improvements in entity scores after adding the retrieval module, which helps the model learn better context that cannot be directly seen from the images (e.g. in some cases GPE and ORG are not clearly present in the images).
And as expected, adding CoLaM to our training pipeline learns better article context, which leads to improvements in entity scores.

\begin{table}[!htbp]
    \centering
    \resizebox{\columnwidth}{!}{
    \begin{tabular}{cccccccccccc}
    \toprule
        & \multirow{2}{*}{Model} &  \multirow{2}{*}{VF} &  \multirow{2}{*}{NF} &  \multirow{2}{*}{RS} & \multirow{2}{*}{CoLaM} & \multicolumn{2}{c}{PERSON} & \multicolumn{2}{c}{GPE} & \multicolumn{2}{c}{ORG} \\
        & & & & & & P & R & P & R & P & R \\
    \hline
        \multirow{6}{*}{\rotatebox{90}{\small GoodNews}}  & $\langle$1$\rangle$ & & & & & 26.58 & 21.99 & 22.80 & 22.34 & 17.84 & 16.42 \\
         & $\langle$2$\rangle$ & \cmark & & & & 28.00 & 24.13 & 24.17 & 24.37 & 19.97 & 19.47 \\
         & $\langle$3$\rangle$ & \cmark & \cmark & & & 29.22 & 25.91 & 24.62 & 24.38 & 21.31 & 20.24 \\
         & $\langle$4$\rangle$ & \cmark & & \cmark & & 29.66 & 24.46 & 24.18 & 24.38 & 21.51 & 21.04 \\
         & $\langle$5$\rangle$ & \cmark & \cmark & \cmark & & 30.33 & 26.32 & 25.33 & 25.17 & 22.10 & 21.11 \\
         & $\langle$6$\rangle$ & \cmark & \cmark & \cmark & \cmark & \textbf{31.00} & \textbf{27.42} & \textbf{25.99} & \textbf{25.62} & \textbf{22.21} & \textbf{21.51} \\
    \midrule
        \multirow{6}{*}{\rotatebox{90}{\scriptsize NYTimes800k}}  & $\langle$1$\rangle$ & & & & & 29.65 & 29.47 & 25.84 & 25.77 & 18.38 & 17.75 \\
         & $\langle$2$\rangle$ & \cmark &  & & & 32.86 & 29.41 & 27.02 & 27.69 & 20.09 & 18.98  \\
         & $\langle$3$\rangle$ & \cmark & \cmark & & & 37.11 & 33.17 & 27.67 & 28.34 & 20.98	& 19.29 \\
         & $\langle$4$\rangle$ & \cmark & & \cmark & & 36.21 & 31.40 & 28.26 & 28.41 & 21.89 & 20.08 \\
         & $\langle$5$\rangle$ & \cmark & \cmark & \cmark  & & 38.53 & 34.22 & 28.28 & 29.00 & 22.66 & 20.42 \\
         & $\langle$6$\rangle$ & \cmark & \cmark & \cmark  & \cmark & 
         \textbf{38.59} & \textbf{35.43} & \textbf{28.44} & \textbf{29.03} & \textbf{23.02} & \textbf{20.86} \\
    \bottomrule
    \end{tabular}
    }
    \caption{Effects of different modules on named entities. Same abbreviation applies as in Table~\ref{tab:ablation_model}.}
    \label{tab:ablation_ent_type}
\end{table}

\noindent \textbf{Evaluation on Different Subsets of the Test Data}

Because the design of our face naming module is guided by the face-name co-occurrence patterns found in News Image Captioning dataset, we split the dataset into three mutually exclusive subsets\footnote{The distribution of samples on the full datasets can be seen in Appendix~\ref{appendix:stat}.
There are no samples without names in the caption that have faces in the corresponding image.} to explore the effectiveness of our face naming module:
1. F\xmark,N\xmark subset with no faces in images and no names in the captions;  2. F\xmark,N\cmark subset with no faces in the images, but has names in the captions and 3. F\cmark,N\cmark subset with faces in the images and names in the captions. We show the results of our model, with or without face naming, on the NYTimes800k dataset in Table \ref{tab:ablation_subsets}.

\begin{table}[!htbp]
    \centering
    \resizebox{\columnwidth}{!}{%
    \begin{tabular}{cccccccc}
    \toprule
       \multirow{2}{*}{Model} & \multirow{2}{*}{Subset} & \multicolumn{2}{c}{PERSON} & \multicolumn{2}{c}{GPE} & \multicolumn{2}{c}{ORG} \\
        & & P & R & P & R & P & R \\
    \hline
          $\langle$2$\rangle$ & F\xmark,N\xmark  & $\diagup$ & $\diagup$ & \textbf{28.71} & 28.75 & 21.56 & 19.84 \\
          $\langle$3$\rangle$ & F\xmark,N\xmark  & $\diagup$ & $\diagup$ & 28.66 & \textbf{29.54} & \textbf{23.43} & \textbf{20.78} \\
    \hline
          $\langle$2$\rangle$ & F\xmark,N\cmark & 30.11 & \textbf{18.94} & 22.05 & 23.17 & \textbf{15.03} & \textbf{16.16} \\
          $\langle$3$\rangle$ & F\xmark,N\cmark  & \textbf{35.43} & 16.65 & \textbf{24.05} & \textbf{23.46} & 14.30 & 14.99 \\
    \hline
          $\langle$2$\rangle$ & F\cmark,N\cmark & \textbf{42.34} & 31.24 & 26.36 & 27.53 & 19.96 & 18.78 \\
          $\langle$3$\rangle$ & F\cmark,N\cmark & 41.77 & \textbf{36.06} & \textbf{27.42} & \textbf{28.11} & \textbf{20.37} & \textbf{18.83} \\
    \hline
    \hline
    \multirow{2}{*}{Model} & \multirow{2}{*}{Subset} & \multicolumn{4}{c}{Caption Generation} & \multicolumn{2}{c}{Named Entites} \\
        & & B & M & R & C & P & R \\
    \hline
          $\langle$2$\rangle$ & F\xmark,N\xmark  & 5.44 & 8.93 & 17.11 & 41.16 & 17.99 & 20.57 \\
          $\langle$3$\rangle$ & F\xmark,N\xmark  & \textbf{5.45} & \textbf{9.04} & \textbf{17.54} & \textbf{43.11} & \textbf{19.91} & \textbf{20.92} \\
    \hline
          $\langle$2$\rangle$ & F\xmark,N\cmark & \textbf{5.64} & \textbf{8.75} & \textbf{17.74} & \textbf{47.03} & 20.91 & \textbf{17.11} \\
          $\langle$3$\rangle$ & F\xmark,N\cmark  & 4.84 & 8.14 & 16.39 & 41.04 & \textbf{22.03} & 16.33 \\
    \hline
          $\langle$2$\rangle$ & F\cmark,N\cmark & 7.72 & 11.18 & 22.11 & 64.95 & 28.44 & 24.39 \\
          $\langle$3$\rangle$ & F\cmark,N\cmark  & \textbf{8.68} & \textbf{12.15} & \textbf{24.16} & \textbf{73.75} & \textbf{29.46} & \textbf{26.55} \\
    \bottomrule
    \end{tabular}
    }
    \caption{Effects of the face naming module on different subsets of NYTimes800k (e.g. F\xmark,N\xmark: subset with no faces in images and no names in captions. See text for details.). Model $\langle$2$\rangle$: $\text{BART}_{\text{base}}$ + visual features; Model $\langle$3$\rangle$: $\langle$2$\rangle$ + face naming. }
    \label{tab:ablation_subsets}
\end{table}

Because the design of the prefix-augmented self attention in our face naming module provides a strong signal of face-name co-occurrence, 
we expect that our model with face naming (Model $\langle$3$\rangle$) would perform much better on PERSON-type entities than the model with only the visual features (Model $\langle$2$\rangle$) on F\cmark,N\cmark subset, and much better overall on both F\xmark,N\xmark and F\cmark,N\cmark subsets.
As shown in Table~\ref{tab:ablation_subsets}, 
on F\cmark,N\cmark subset, 
we increase the recall from 31.24 to 36.06,
while maintaining the same level of precision after adding the face naming module into Model $\langle$2$\rangle$. And the CIDEr scores of the generated captions from Model $\langle$3$\rangle$ are significantly higher than the counterpart.
On top of that, the correctness of the generated entities overall improves on both F\xmark,N\xmark and F\cmark,N\cmark subsets.

The F\xmark,N\cmark subset contains samples without face-name co-occurrence pattern as we modeled in our face naming module. 
It covers around 11\% data in each dataset, as shown in Table 2 in the appendix.
It can be seen from Table \ref{tab:ablation_subsets}, the trade-off of significantly increasing the model performance on F\xmark,N\xmark and F\cmark,N\cmark subsets is that the model would generate worse captions on the F\xmark,N\cmark subset. Model $\langle$3$\rangle$ performs worse on most of the evaluation metrics than Model $\langle$2$\rangle$. 
However, one interesting finding is that Model $\langle$3$\rangle$ achieves a much higher precision 
while reaching a lower recall 
with PERSON-type entity names on F\xmark,N\cmark subset. 
Meanwhile, an opposite trend can be found on F\cmark,N\cmark subset. 
With the face naming module, our model tends to generate less PERSON-type entity names when there is no face, and to generate more PERSON-type entity names
otherwise, showing its effectiveness.
The same ablation study on GoodNews dataset is presented in Appendix~\ref{appendix:more_ablation}.

\section{Conclusion}

In this paper, we introduce a new framework for utilizing visual inputs in News Image Captioning. 
Inspired by human attention mechanisms, we developed a face naming module for aligning names with faces in images, based on face-name co-occurrence patterns. 
For context that cannot be visually grounded in the images, we utilize CLIP for sentence retrieval from articles, aiding comprehension. 
To address the imbalance between article and image context in captions, we introduce CoLaM, guiding the model to focus more on article content. Our extensive experiments demonstrate the effectiveness of our method, which achieves more than 6-point improvement in CIDEr scores over the previous state-of-the-art on two commonly-used News Image Captioning datasets.

\section{Limitations}

Our face naming module effectively aligns faces in images to names in articles/captions, which can be directly visually grounded from the images and trigger higher human attention priority.
However, for contexts like time or organizations that typically cannot be directly visually grounded, we depend on CLIP retrieval to infer links between articles and images. A potential improvement involves designing specific modules for these types of contexts. Additionally, our CoLaM approach currently treats all image-caption-article triplets equally, applying the same constraints during training. A valuable area for future research would be to investigate a weighting mechanism that selectively adjusts the margin loss computation for these triplets.

\section{Acknowledgement}

This work is funded by the Flanders AI Research Program and the China Scholarship Council.

\bibliography{anthology,custom}

\appendix

\section{Overview of the Appendix}
\label{appendix:summary}

In the appendix, we first provide more implementation details and dataset statistics in Appendix~\ref{appendix:implementation} and \ref{appendix:stat}, respectively.
This is followed by human evaluation of the generated captions form our method in Appendix~\ref{appendix:human}.
Following that, to showcase the connection and difference between News Image Captioning and Text Summarization, we conduct additional experiments on text summarization as presented in Appendix~\ref{appendix:text_summ}.
Further, we provide more ablation studies on different subsets of the GoodNews test data in Appendix~\ref{appendix:more_ablation}. 
After that, we show experiments on comparing our method to large vision-language models in Appendix~\ref{appendix:large_vlms}.
Then we provide a in-depth study of our CoLaM in Appendix~\ref{appendix:colam}.
We conclude the appendix with qualitative analyses on the generated captions in Appendix~\ref{appendix:qualitative}.

\section{More Implementation Details}
\label{appendix:implementation}

Extra Data regard the external data and include:
JoGANIC~\cite{yang-etal-2021-journalistic}: Wikipedia database; DiscExt CapGen~\cite{ZHANG2022108429}: 2.755 million caption-style pairs. 
\citet{rajakumar-kalarani-etal-2023-lets}: 1.2 million paired News Image Captioning data from VisualNews~\cite{liu-etal-2021-visual}.
$\bullet$: Apart from the language model\&vision backbones, Focus!~\cite{zhou-etal-2022-focus} uses CLIP and domain specific relation extraction model OpenNRE~\cite{han-etal-2019-opennre} for context retrieval.

The MLP network in visual feature generation module consists of two linear layers with hyperbolic tangent activation in between. The two linear layers are of shape Linear($dim_{model}$, $dim_{model}\times10$) and Linear($dim_{model}\times10$, $dim_{model}\times20$), where  $dim_{model}$ is the model dimension of the BART backbone. We reshape the mapped visual feature from $(batch_{size}, dim_{model}\times20)$ to  $(batch_{size}, 20, dim_{model})$.

To obtain the name embedding in our face naming module, the same embedding layer structure as in BART is adopted.
Given a chain like ”name1 $\langle$ENT$\rangle$ name2 . . .”, with $\langle$ENT$\rangle$ being the added token in vocabulary,we compute the word embedding $H_N$ of the chain. 
Then we first limit the maximum length of the chain of person names in the articles to be 80 tokens. Since the names are taken from articles with maximum 512 tokens, by our estimation, 80 is enough to cover all the names in the article segments. The feed forward layers succeeding PA Self Attention in Figure~\ref{fig:model} (b) map the hidden states of names from $(batch_{size}, 80, dim_{model})$ to $(batch_{size}, 20, dim_{model})$.


We use the transformers package to build our models.
For $\text{BART}_{\text{base}}$, we adopt the "facebook/bart-base" checkpoint; while for  $\text{BART}_{\text{large}}$, we adopt the "patrickvonplaten/bart-large-fp32" checkpoint.
The default vocabulary size is 50265. 
The training of $\text{Ours}_{\text{base}}$ and $\text{Ours}_{\text{large}}$ takes roughly 1 and 2 days on 1$\times$A100, respectively.
For the GoodNews dataset, the full articles are used. While for NYTimes800k, we follow the standard protocol \cite{Tran_2020_CVPR} to use the 512 tokens surrounding the images. 
For our full model, we apply length penalty of 2 during decoding.
Following \citet{Tran_2020_CVPR}, we adopt pycocoevalcap package and spacy package (ver. 2.1.9) for evaluating generated captions and entity scores, respectively.

\section{Dataset Statistics}
\label{appendix:stat}

In this section, we provide dataset statistics of GoodNews~\cite{Biten_2019_CVPR} and NYTimes800k~\cite{Tran_2020_CVPR}. The overall statistics of two datasets are provided in Table~\ref{tab:data_overall}. 

\begin{table}[!htbp]
    \begin{center}
    \resizebox{\columnwidth}{!}{
    \begin{tabular}{ccc}
        \toprule
        & GoodNews & NYTimes800k \\
        \hline
        Number of images & 462642 & 792971 \\
        Average article length &  451  & 974 \\
        Average caption length &  18  &  18 \\
        \hline
        \% of captions with named entities &  97\%  &  96\% \\
        \% of captions with person names &  68\%  &  68\% \\
        \% of images with faces &  56\%  &  57\% \\
        \bottomrule
    \end{tabular}
    }
    \end{center}
    \caption{Dataset statistics for GoodNews and NYTimes800k}
    \label{tab:data_overall}
\end{table}

Table~\ref{tab:face-name} presents the statistics of face-name co-occurrence patterns in two datasets.

\begin{table}[!htbp]
\begin{center}
\resizebox{\columnwidth}{!}{
\begin{tabular}{lcccc}
\toprule
Dataset & F \cmark, N \cmark & F \xmark, N \xmark & F \cmark, N \xmark & F \xmark, N \cmark\\
\hline
GoodNews  & 56.30\% & 31.91\% & 0\% & 11.79\% \\
NYTimes800k & 56.91\% & 32.05\% & 0\% & 11.04\% \\
\bottomrule
\end{tabular}
}
\end{center}
\caption{Statistics of face-name co-occurrence patterns in two 
News Image Captioning datasets.\protect \footnotemark "F \cmark, N \xmark" refers to samples with faces in images, but no names in captions.}
\label{tab:face-name}
\end{table}

\footnotetext{Calculation based on face features provided in the datasets.}

\section{Human Evaluation}
\label{appendix:human}

We present human evaluation in Table~\ref{tab:human}. We hire three graduate students with domain knowledge 
to rank 50 randomly sampled captions on correctness and fluency 
from 1 to 5, and pick their preferred caption.
Captions generated by our method better align with human judgement (\textit{C}=3.83), and are preferred by humans in \textbf{67\%} of the cases.

\begin{table}[!htbp]
    \centering
    \resizebox{0.8\columnwidth}{!}{
    \begin{tabular}{lcc|c}
    \toprule
        Model & Correctness(\textit{C}) & Fluency(\textit{F}) & Preferred by \\
    \hline
        Baseline  & $3.15_{\pm0.13}$ & $4.67_{\pm0.25}$ & $16\%_{\pm4\%}$ \\
        Ours & $\textbf{3.83}_{\pm0.19}$ & $\textbf{4.86}_{\pm0.10}$ & $\textbf{67\%}_{\pm5\%}$ \\
    \bottomrule
    \end{tabular}
    }
    \caption{Human evaluation for generated captions.}
    \label{tab:human}
\end{table}

\section{Additional Experiments on Text Summarization}
\label{appendix:text_summ}

Since our work is also closely related to text summarization, in this section, we present more experiments on text summarization with BART.

\subsection{Experiments with frozen BART}
By keeping the BART backbone frozen during training, we can have a better idea of whether the modules we designed can guide the caption generation effectively. 
We present the results in Table~\ref{tab:freeze}. The frozen $\text{BART}_{\text{base}}$ works poorly by only achieving 6.56 CIDEr score. While after adding our modules, even without training the BART backbone, we achieve fairly good results with CIDEr=56.66. This significant improvement in performance shows that our added modules can effectively guide the generation process of BART.

\begin{table}[!htbp]
        \centering
        \resizebox{\columnwidth}{!}{
		\begin{tabular}{lcccc}
        \toprule
        Model & BLEU-4 &  METEOR & ROUGE-L & CIDEr \\
        \hline
        Frozen $\text{BART}_{\text{base}}$ (pure text)  & 1.68 & 8.33 & 11.43 & 6.56 \\
        Ours + frozen $\text{BART}_{\text{base}}$ & 5.85 & 9.93 & 20.86 & 56.66 \\
        \bottomrule
        \end{tabular}
        }
        \caption{Performance comparison with frozen BART on GoodNews dataset}
        \label{tab:freeze}
\end{table}

\subsection{Experiments with summarization with retrieval}

Our retrieval method aims to locate the sentences that are semantically closer to the images. Without adding any visual information into the model, we perform text summarization on the retrieved segments only as shown in Table~\ref{tab:text}. The large improvements in performance (CIDEr=55.24 $\rightarrow$ 59.21) by changing the inputs from the full articles to the retrieved segments prove that our retrieval component can locate more accurate semantic information from the articles.

\begin{table}[!htbp]
\begin{center}
\resizebox{\columnwidth}{!}{%
\begin{tabular}{lcccc}
\toprule
Model & BLEU-4 &  METEOR & ROUGE-L & CIDEr \\
\hline
$\text{BART}_{\text{base}}$ & 6.14 & 9.69 & 19.51 & 55.24 \\
$\text{BART}_{\text{base}}$ + our retrieval & 6.43 & 10.03 & 20.50 & 59.21 \\
\bottomrule
\end{tabular}
}
\end{center}
\caption{Text summarization with $\text{BART}_{\text{base}}$ on GoodNews with our retrieved article segments.}
\label{tab:text}
\end{table}

\section{Additional Ablation Studies}
\label{appendix:more_ablation}

\noindent \textbf{Ablation Study on Number of Retrieved Sentences}

We evaluate the impact of the number of retrieved sentences, as outlined in Table \ref{tab:ablation_num_sent}.
Here we do not apply CoLaM to show a clear image of how the number of retrieved sentences can affect the performance of our model.
Our results indicate consistent performance across the range of 7-10 retrieved sentences on both GoodNews and NYTimes800k datasets.
It's worth noting that while the top CIDEr score doesn't consistently align with the highest achievements in other evaluation metrics, such as BLEU-4 (7.02) and METEOR (10.77) which are attained with retrieving 9 sentences in the case of GoodNews, models with the highest CIDEr score generally maintain strong performance in other metrics.

\begin{table}[!htbp]
    \centering
    \resizebox{\columnwidth}{!}{
    \begin{tabular}{cccccccc}
    \toprule
        & \multirow{2}{*}{\# of sent} & \multicolumn{4}{c}{Caption Generation} & \multicolumn{2}{c}{Named Entites} \\
        &  & B & M & R & C & P & R \\
    \hline
        \multirow{4}{*}{\rotatebox{90}{\footnotesize GoodNews}}  & 7 & 6.92 & 10.74 & 21.68 & 62.62 & 23.39 & 21.27 \\
         & 8 & 7.00 & 10.75 & \textbf{21.79} & \textbf{64.07} & \textbf{23.69} & \textbf{21.50}  \\
         & 9 & \textbf{7.02} & \textbf{10.77} & 21.61 & 62.64 & 23.19 & 21.30 \\
         & 10 & 6.83 & 10.66 & 21.64 & 63.43 & 23.49 & 21.21 \\
    \midrule
        \multirow{4}{*}{\rotatebox{90}{\scriptsize NYTimes800k}} & 7 & 7.38 & 10.90 & 21.61 & 63.13 & 26.62 & 24.39 \\
         & 8 & 7.43 & 10.92 & 21.52 & 62.55 & 26.85 & 24.28 \\
         & 9 & 7.50 & \textbf{10.98} & 21.53 & 62.84 & 26.58 & 24.60 \\
         & 10 & \textbf{7.53} & \textbf{10.98} & \textbf{21.63} & \textbf{63.95} & \textbf{26.94} & \textbf{24.72} \\
    \bottomrule
    \end{tabular}
    }
    \caption{Influence of the number of retrieved sentences}
    \label{tab:ablation_num_sent}
\end{table}

\noindent \textbf{Ablation on Different Subsets of the Test Data (GoodNews)}

We present the ablation study on different subsets of the GoodNews test data in Table \ref{tab:ablation_subsets_goodnews}. For F\xmark,N\xmark and F\cmark,N\cmark subsets, we observe similar trend in performance improvements when adding entity prefix into the model as the case for the subsets of GoodNews test data. The biggest improvements in entity scores can be seen in recall for PERSON type entities on F\cmark,N\cmark subset when we add entity prefix into the model (25.15$\rightarrow$27.77). And the quality of the generated captions is drastically enhanced on F\cmark,N\cmark subset, as demonstrated by an approximate 7.5-percentage-point improvements (from 68.11 to 73.24).

Interestingly, on F\xmark,N\cmark subset of GoodNews test data, we observe uniformly decreasing in all metrics when adding name features to the model. While on on F\xmark,N\cmark subset of NYTimes800k test data, we observe a small improvement in entity precision.
Moreover, a slightly different pattern in entity scores can be observed on the F\xmark,N\cmark subset from two datasets.
It shows that for subsets without the face-name co-occurrence pattern we modeled, the performance of our model is somewhat dependent to the data distribution. 
Notably, the F\xmark,N\cmark subset constitutes approximately 11\% of the entire dataset. Consequently, the substantial performance improvements observed in the remaining 89\% of the data contribute to generating superior captions in the aggregate.

\begin{table}[!htbp]
    \centering
    \resizebox{\columnwidth}{!}{%
    \begin{tabular}{cccccccc}
    \toprule
       \multirow{2}{*}{Model} & \multirow{2}{*}{Subset} & \multicolumn{2}{c}{PERSON} & \multicolumn{2}{c}{GPE} & \multicolumn{2}{c}{ORG} \\
        & & P & R & P & R & P & R \\
    \hline
          (2) & F\xmark,N\xmark  & $\diagup$ & $\diagup$ & 25.85 & 26.67 & 20.58 & 18.91 \\
          (3) & F\xmark,N\xmark  & $\diagup$ & $\diagup$ &  \textbf{26.43} & \textbf{27.51} & \textbf{21.95} & \textbf{19.39} \\
    \hline
          (2) & F\xmark,N\cmark & 26.90 & \textbf{18.03} & 18.18 & \textbf{18.99} & 14.11 & 14.93 \\
          (3) & F\xmark,N\cmark  & \textbf{29.83} & 14.88 & \textbf{18.52} & 18.22 & \textbf{16.02} & \textbf{16.28}  \\
    \hline
          (2) & F\cmark,N\cmark & \textbf{33.54} & 25.15 & 24.03 & \textbf{23.60} & 20.71 & 20.63 \\
          (3) & F\cmark,N\cmark  & 31.50 & \textbf{27.77} & \textbf{24.31} & 23.10 & \textbf{21.91} & \textbf{21.48} \\
    \hline
    \hline
    \multirow{2}{*}{Model} & \multirow{2}{*}{Subset} & \multicolumn{4}{c}{Caption Generation} & \multicolumn{2}{c}{Named Entites} \\
        & & B & M & R & C & P & R \\
    \hline
          $\langle$2$\rangle$ & F\xmark,N\xmark  & 5.48 & 9.10 & 17.70 & 44.30 & 17.96 & 19.92 \\
          $\langle$3$\rangle$ & F\xmark,N\xmark  & \textbf{5.58} & \textbf{9.34} & \textbf{18.42} & \textbf{46.66} & \textbf{19.69} & \textbf{20.05} \\
    \hline
          $\langle$2$\rangle$ & F\xmark,N\cmark & \textbf{5.48} & \textbf{9.00} & \textbf{18.18} & \textbf{48.02} & \textbf{19.08} & \textbf{15.89} \\
          $\langle$3$\rangle$ & F\xmark,N\cmark  & 4.81 & 8.48 & 17.47 & 44.28 & 18.96 & 14.61 \\
    \hline
          $\langle$2$\rangle$ & F\cmark,N\cmark & 7.35 & 10.96 & 22.26 & 68.11 & 24.79 & 21.54 \\
          $\langle$3$\rangle$ & F\cmark,N\cmark  & \textbf{7.82} & \textbf{11.58} & \textbf{23.49} & \textbf{73.24} & \textbf{24.82} & \textbf{22.62} \\     
    \bottomrule
    \end{tabular}
    }
    \caption{Effectiveness of face naming module with $\text{Ours}_{\text{base}}$ on different subsets of GoodNews. Model $\langle$2$\rangle$: $\text{BART}_{\text{base}}$ + visual features; Model $\langle$3$\rangle$: $\langle$2$\rangle$ + name features from face naming module. F\xmark,N\xmark: subset with no faces in images and no names in captions. }
    \label{tab:ablation_subsets_goodnews}
\end{table}

\noindent \textbf{Ablation on Lead3 Sentences and Varying Feature Length}

We present the ablation study on Lead3 sentences in CLIP retrieval and varying feature length in Table~\ref{tab:lead3_length}.
As expected, removing Lead3 sentences from the retrieved segments harms the performance of our method, due to lack of global context from the articles. And different feature lengths yield similar performance, while feature length=20 achieves the best CIDEr score.

\begin{table}[!htbp]
    \centering
    \resizebox{\columnwidth}{!}{
    \begin{tabular}{cllcccc}
    \toprule
         & Lead3 & Length &  BLEU-4 &  METEOR  & ROUGE-L & CIDEr \\
    \hline
        \multirow{5}{*}{\rotatebox{90}{GoodNews}}  & \xmark  & 20 & 6.80 & 10.67 & 21.43 & 61.37 \\
         & \cmark(Ours) & 20 & 7.00 & 10.75 & 21.79 & 64.07 \\
        \cline{2-7}
         & \cmark & 16  & 6.90 & 10.70 & 21.50 & 62.04 \\
         & \cmark & 20(Ours) & 7.00 & 10.75 & 21.79 & 64.07 \\
         & \cmark & 24 & 6.88 & 10.65 & 21.45 & 61.86 \\
    \bottomrule
    \end{tabular}
    }
    \caption{Performance comparison w/ or w/o Lead3  \& w/ varying feature length using $\text{Ours}_{\text{base}}$ w/o CoLaM}
    \label{tab:lead3_length}
\end{table}

\noindent \textbf{Additional Ablation on Face Naming Module}

Table~\ref{tab:nomap} shows results on replacing the face naming module with feature concatenation (w/ retrieved segments (RS)).
As expected, by replacing our face naming module with simple concatenation features, we observe significant degradation in performance.

\begin{table}[!htbp]
    \centering
    \resizebox{\columnwidth}{!}{
    \begin{tabular}{lccllll}
    \toprule
     Dataset & RS & Feature Integration &  BLEU-4 &  METEOR  & ROUGE-L & CIDEr \\
    \hline
        GoodNews & \cmark & concatenation  & 6.76 & 10.46 & 21.02 & 60.87 \\
        GoodNews & \cmark & face naming module & \textbf{7.00} & \textbf{10.75} & \textbf{21.79} & \textbf{64.07}  \\
    \bottomrule
    \end{tabular}
    }
    \caption{Ablation study on face naming module and feature concatenation ($\text{Ours}_{\text{base}}$ w/o CoLaM).}
    \label{tab:nomap}
\end{table}

\section{Experiments with Large Vision-Language Models}
\label{appendix:large_vlms}

In this section, we conduct experiments using large vision-language models (VLMs) on two News Image Captioning datasets.
We select two of the most performant large VLMs, namely InstructBLIP~\cite{dai2023instructblip} and LLaVA-1.5~\cite{liu2023improved}.
We adopt the Vicuna7B version of both models.

\noindent \textbf{Comparing to Large VLMs with Prompt}

We first report experimental results using frozen large VLMs with carefully designed prompt as:

\begin{itemize}
    \item For InstructBLIP, we prompt the model using "News article:<article> Generate news image caption:".
    \item For LLaVA-1.5, we prompt the model using "USER: <image>\textbackslash nNews article: <article> Generate news image caption:\textbackslash nASSISTANT:".
\end{itemize}
where <article> indicates the news article text, and <image> denotes the news image.

\begin{table}[!htbp]
    \centering
    \resizebox{\columnwidth}{!}{
    \begin{tabular}{lcllll}
    \toprule
     Dataset & Method &  BLEU-4 &  METEOR  & ROUGE-L & CIDEr \\
    \hline
        GoodNews & LLaVA-1.5 & 2.32 & \underline{6.92} & \underline{12.67} & 16.21 \\
        GoodNews & InstructBLIP & \underline{2.41} & 5.88 & 10.66 & \underline{17.50}  \\
        GoodNews & $\text{Ours}_{\text{large}}$ & \textbf{8.60} & \textbf{12.39} & \textbf{23.38} & \textbf{71.96} \\
        \hline
        NYTimes800k & LLaVA-1.5 & 2.41 & \underline{7.40} & \underline{12.52} & 14.83 \\
        NYTimes800k & InstructBLIP & \underline{3.09} & 7.14 & 12.45 & \underline{19.17}  \\
        NYTimes800k & $\text{Ours}_{\text{large}}$ & \textbf{9.24} & \textbf{12.57} & \textbf{23.44} & \textbf{71.65} \\
    \bottomrule
    \end{tabular}
    }
    \caption{Results of $\text{Ours}_{\text{large}}$ and large VLMs (LLaVA-1.5 and InstructBLIP) on two News Image Captioning datasets. The large VLMs are frozen, but are prompted with our specifically designed text prompts.}
    \label{tab:vlm_prompt}
\end{table}

The results are presented in Table~\ref{tab:vlm_prompt}. 
It is evident that the task remains challenging even for large VLMs.

\noindent \textbf{Comparing to Fine-tuned Large VLMs}

We also compare our method to the fully fine-tuned large VLMs as presented in Table~\ref{tab:vlm_finetune}.
Despite having a language model with roughly 5\% parameters compared to large VLMs ($\text{BART}_{\text{large}}$ (400M) vs Vicuna (7B)), 
our method yields comparable or better performance on both datasets.
It underscores the significance of our method, which is not only lightweight but also demonstrates exceptional performance, thereby fulfilling a crucial need in the community.

\begin{table}[!htbp]
    \centering
    \resizebox{\columnwidth}{!}{
    \begin{tabular}{lclll}
    \toprule
     Dataset & Method &  BLEU-4  & ROUGE-L & CIDEr \\
    \hline
        GoodNews & $\text{LLaVA-1.5}^\dag$ & 7.04 & \underline{24.37} & \underline{73.52} \\
        GoodNews & $\text{InstructBLIP}^\dag$ & \textbf{9.53} & \textbf{25.61} & \textbf{78.03}  \\
        GoodNews & $\text{Ours}_{\text{large}}$ & \underline{8.60} & 23.38 & 71.96 \\
        \hline
        NYTimes800k & $\text{LLaVA-1.5}^\dag$  & 6.06 & 22.80 & 62.41 \\
        NYTimes800k & $\text{InstructBLIP}^\dag$ & \textbf{10.05} & \textbf{25.45} & \textbf{75.95}  \\
        NYTimes800k & $\text{Ours}_{\text{large}}$ & \underline{9.24} & \underline{23.44} & \underline{71.65} \\
    \bottomrule
    \end{tabular}
    }
    \caption{Results of $\text{Ours}_{\text{large}}$ and fully fine-tuned large VLMs (LLaVA-1.5 and InstructBLIP) on two News Image Captioning datasets. $\dag$: The results of the fine-tuned large VLMs are directly taken from~\cite{zhang2024entity}}
    \label{tab:vlm_finetune}
\end{table}

\section{In-depth Study of CoLaM}
\label{appendix:colam}

We conduct ablation studies of CoLaM using $\text{BART}_{\text{base}}$ as the LM backbone. Limited to resources, the batch size is set to 24 for all ablation studies in this section.

\noindent \textbf{Impact of the Margin Values $\Delta$}

We present the results with varying values for the margin parameter $\Delta$ in Table~\ref{tab:margin}. 
Since the range of cosine similarity is within $[-1,1]$, with $\Delta=1.0$, the optimization of CoLaM affects all samples in the datasets. 
Our model mainly promote the visual inputs during generation, which makes the consistently added constraint from our CoLaM more favorable.
As expected, the model reaches the best performance when we set $\Delta=1.0$.

\begin{table}[!htbp]
\centering
\resizebox{0.85\columnwidth}{!}{
\begin{tabular}{clccccccccc}
\toprule
& \multirow{2}{*}{$\Delta$} & \multirow{2}{*}{$\alpha$} & \multicolumn{4}{c}{Caption Generation} & \multicolumn{2}{c}{Named Entities} \\ 

& & & B &  M & R & C & P & R  \\ \hline
\multirow{5}{*}{\rotatebox{90}{GoodNews}} 
& \xmark & \xmark & \underline{6.93} & \underline{10.75} & \underline{21.69} & \underline{62.94} & \underline{23.41} & \underline{21.46} \\
& 0.4 & 1.0 & 6.95 & 10.75 & 21.74 & 63.87 & 23.38 & 21.45 \\
& 0.6 & 1.0 & 7.00 & 10.81 & 21.73 & 63.48 & 23.25 & 21.39 \\
& 0.8 & 1.0 & 7.14 & 10.90 & 21.75 & 63.65 & 23.19 & 21.75 \\
& 1.0 & 1.0 & \textbf{7.19} & \textbf{10.94} & \textbf{21.96} & \textbf{65.06} & \textbf{23.78} & \textbf{21.81} \\
\hline
\multirow{4}{*}{\rotatebox{90}{NYTimes800k}} 
& \xmark & \xmark & \underline{7.63} & \underline{11.00} & \underline{21.40} & \underline{62.03} & \underline{25.44} & \underline{23.74} \\
& 0.4 & 1.0 & 7.59 & 11.00 & 21.35 & 62.76 & 26.14 & 24.48 \\
& 0.6 & 1.0 & 7.66 & 11.02 & 21.52 & 63.09 & 26.46 & 24.57 \\
& 0.8 & 1.0 & 7.52 & 10.96 & 21.48 & 62.90 & \textbf{26.55} & 24.62 \\
& 1.0 & 1.0 & \textbf{7.73} & \textbf{11.14} & \textbf{21.66} & \textbf{63.44} & 26.40 & \textbf{24.84} \\
\bottomrule

\end{tabular}
}
\caption{Impact of the choice of the margin ($\Delta$) on the performance. \xmark: model trained without CoLaM.}
\label{tab:margin}
\end{table}

\noindent \textbf{Impact of the Loss Weights $\alpha$}

Table~\ref{tab:alpha} shows the impact of the weight $\alpha$ for $\mathcal{L}_m$ in CoLaM.
We obtain similar results with different values of $\alpha$, showing that CoLaM is less sensitive to the weights.
Setting $\alpha$ to 1.0 or 2.0 yields similar performance.
In practice, we suggest to select $\alpha=1.0$ to avoid unnecessary hyperparameter tuning.

\begin{table}[h!]
\centering
\resizebox{0.83\columnwidth}{!}{
\begin{tabular}{clccccccccc}
\toprule
& \multirow{2}{*}{$\alpha$} & \multirow{2}{*}{$\Delta$} & \multicolumn{4}{c}{Caption Generation} & \multicolumn{2}{c}{Named Entities} \\ 

& & & B &  M & R & C & P & R  \\ \hline
\multirow{4}{*}{\rotatebox{90}{GoodNews}} & \xmark & \xmark & \underline{6.93} & \underline{10.75} & \underline{21.69} & \underline{62.94} & \underline{23.41} & \underline{21.46} \\
& 0.5 & 1.0 & 7.15 & 10.90 & 21.87 & 64.54 & 23.51 & 21.79 \\
& 1.0 & 1.0 & 7.19 & 10.94 & 21.96 & \textbf{65.06} & \textbf{23.78} & 21.81 \\
& 2.0 & 1.0 & \textbf{7.27} & \textbf{11.02} & \textbf{21.97} & 64.53 & 23.60 & \textbf{22.13} \\
\hline
\multirow{4}{*}{\rotatebox{90}{\footnotesize NYTimes800k}} &\xmark & \xmark & \underline{7.63} & \underline{11.00} & \underline{21.40} & \underline{62.03} & \underline{26.26} & \underline{24.46} \\
& 0.5 & 1.0 & \textbf{7.79} & \textbf{11.15} & 21.65 & 63.15 & 26.56 & 24.86 \\
& 1.0 & 1.0 & 7.73 & 11.14 & 21.66 & 63.44 & 26.40 & 24.84 \\
& 2.0 & 1.0 & 7.73 & 11.14 & \textbf{21.72} & \textbf{63.78} & \textbf{26.61} & \textbf{24.96} \\
\bottomrule

\end{tabular}
}
\caption{Impact of weights $\alpha$ for our $\mathcal{L}_m$. \xmark: model trained without CoLaM.}
\label{tab:alpha}
\end{table}

\noindent \textbf{Using Encoder or Decoder Hidden States}

Since we learn the multimodal interaction only in the encoder, comparing the performance using the last encoder hidden states and the last decoder hidden states 
provides insights into whether the additional information from the caption influences CoLaM.
As shown in Table~\ref{tab:enc}, we obtain similar results with two types of hidden states, indicating the extra information from the caption does not have a significant impact on the functioning of CoLaM in the training pipeline.

\begin{table}[!htbp]
\centering
\resizebox{\columnwidth}{!}{
\begin{tabular}{llcccccccccc}
\toprule
\multirow{2}{*}{Dataset} &\multirow{2}{*}{Hidden States} & \multirow{2}{*}{$\alpha$} & \multirow{2}{*}{$\Delta$} & \multicolumn{4}{c}{Caption Generation} & \multicolumn{2}{c}{Named Entities} \\ 

& & & & B &  M & R & C & P & R  \\ \hline
GoodNews & Encoder & 1.0 & 1.0 & 7.16 & 10.93 & 22.05 & 65.24 & 23.78 & 21.81 \\
GoodNews & Decoder & 1.0 & 1.0 & 7.19 & 10.94 & 21.96 & 65.06 & 23.78 & 21.81 \\
\hline
NYTimes800k & Encoder & 1.0 & 1.0 & 7.65 & 11.04 & 21.44 & 62.81 & 26.43 & 24.58 \\
NYTimes800k & Decoder & 1.0 & 1.0 & 7.73 & 11.14 & 21.66 & 63.44 & 26.40 & 24.84 \\
\bottomrule

\end{tabular}
}
\caption{Impact of using the last hidden states from the encoder or the decoder.}
\label{tab:enc}
\end{table}



\noindent \textbf{Generalization Ability of CoLaM}

CoLaM presents extraordinary generalization ability to other model architectures. We implemented the prefix-based method as proposed by \citet{zhang-et-al-mm22}, which is the baseline version of the previous SOTA NewsMEP. NewsMEP utilizes visual and entity prefixes to guide the learning of the multimodal language model. Following \citet{zhang-et-al-mm22}, we use the visual prefix to guide the language model, and term this model as $\text{NewsMEP'}_{\text{base}}$  for clarity.

\begin{table}[!htbp]
\centering
\resizebox{\columnwidth}{!}{
\begin{tabular}{llcccccccc}
\toprule
\multirow{2}{*}{Method} & \multirow{2}{*}{Dataset} & \multicolumn{4}{c}{Caption Generation} & \multicolumn{2}{c}{Named Entities} \\ 
& & B &  M & R & C & P & R  \\ \hline
$\text{NewsMEP'}_{\text{base}}$ & GoodNews & 6.45 & 9.99 & 20.15 & 57.78 & 22.45 & 20.47 \\
$\text{NewsMEP'}_{\text{base}}$ + CoLaM & GoodNews & \textbf{6.73} & \textbf{10.42} & \textbf{20.86} & \textbf{60.63} & \textbf{22.89} & \textbf{21.05} \\
\hline
$\text{NewsMEP'}_{\text{base}}$ & NYTimes800k & 6.56 & 9.95 & 19.50 & 53.36 & 23.79 & 22.09 \\
 $\text{NewsMEP'}_{\text{base}}$ + CoLaM & NYTimes800k & \textbf{7.20} & \textbf{10.48} & \textbf{20.54} & \textbf{57.31} & \textbf{24.86} & \textbf{23.72} \\
 \bottomrule
\end{tabular}
}
\caption{Results for integrating CoLaM into the training pipeline of $\text{NewsMEP'}_{\text{base}}$}
\label{tab:colam_newsmep}
\end{table}

Just as the experiments with our model, without changing any training or architectural designs of $\text{NewsMEP'}_{\text{base}}$ and by simply adding our CoLaM to its training pipeline, we obtain significant performance gain over the original $\text{NewsMEP'}_{\text{base}}$. It shows that our CoLaM can be a valuable addition to the field, and possibly the standard method in any other News Image Captioning model’s training pipeline in the future.

\section{Qualitative Analysis}
\label{appendix:qualitative}

\subsection{Qualitative Examples without CoLaM}

Table \ref{tab:qualitative} shows two examples of generated captions. In the first one, 
our proposed model generates a caption that matches the ground truth with the exception of a missing quote. In the second one, models with face naming module manage to capture important context from the news article, while after adding the sentence retrieval component, more precise context is generated (e.g. Mayor Nan Whaley of Dayton), showing the merit of our method.

\begin{figure}[!htbp]
    \centering
    \includegraphics[width=\columnwidth]{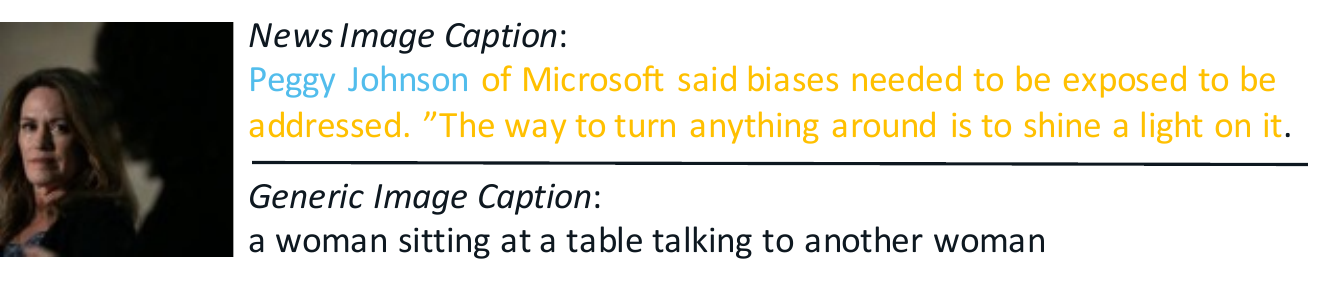}
    \caption{
    Comparison between two types of image captions for image in Table~\ref{tab:qualitative} (1) }
    \label{fig:example_image_appendix1}
\end{figure}

\begin{figure}[!htbp]
    \centering
    \includegraphics[width=\columnwidth]{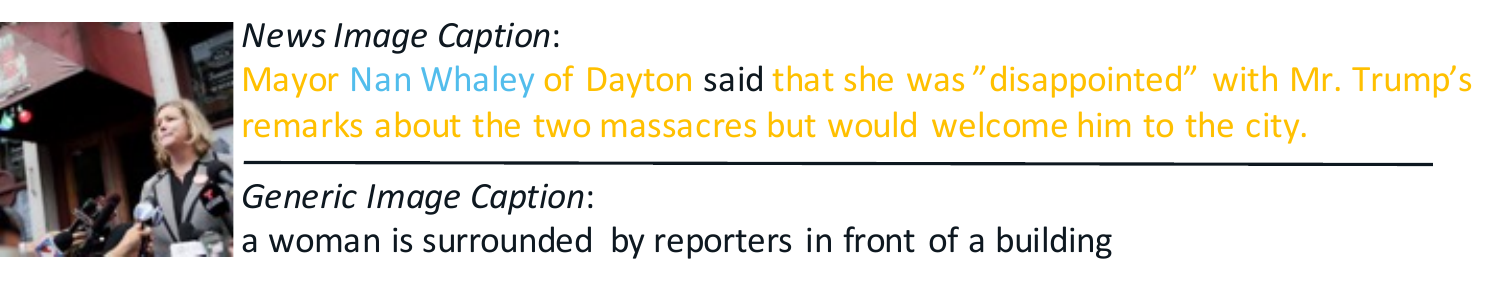}
    \caption{
    Comparison between two types of image captions for image in Table~\ref{tab:qualitative} (2) }
    \label{fig:example_image_appendix2}
\end{figure}

For reference, we also provide the example generic image captions for news images in Table~\ref{tab:qualitative} in Figures~\ref{fig:example_image_appendix1} and \ref{fig:example_image_appendix2}.

\begin{figure}[!htbp]
\centering
\hdashrule[0.5ex][x]{\linewidth}{0.5pt}{1.5mm}
\includegraphics[width=\columnwidth]{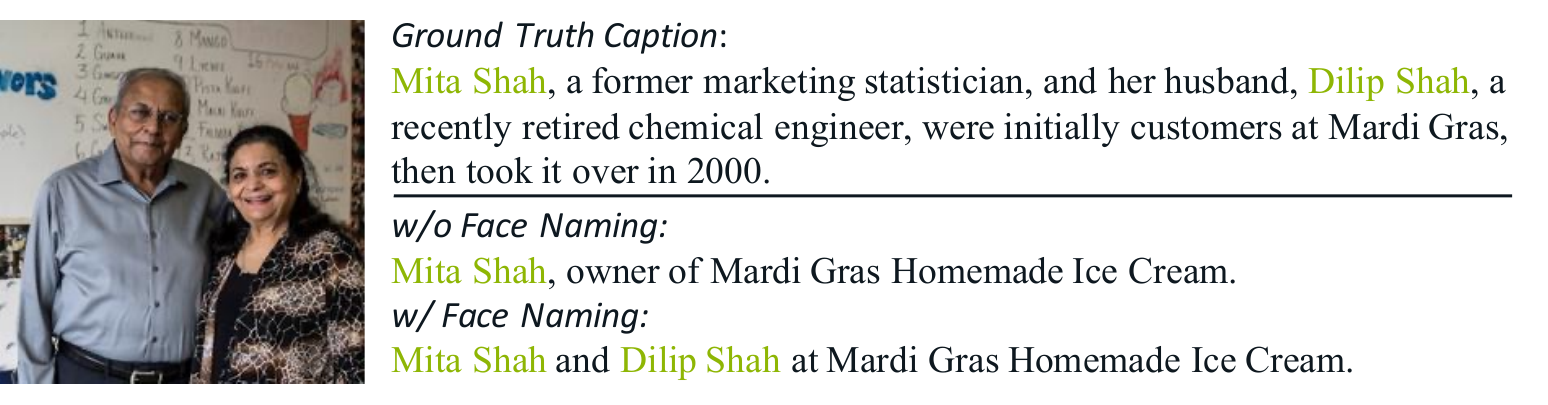}
\hdashrule[0.5ex][x]{\linewidth}{0.5pt}{1.5mm}
\includegraphics[width=\columnwidth]{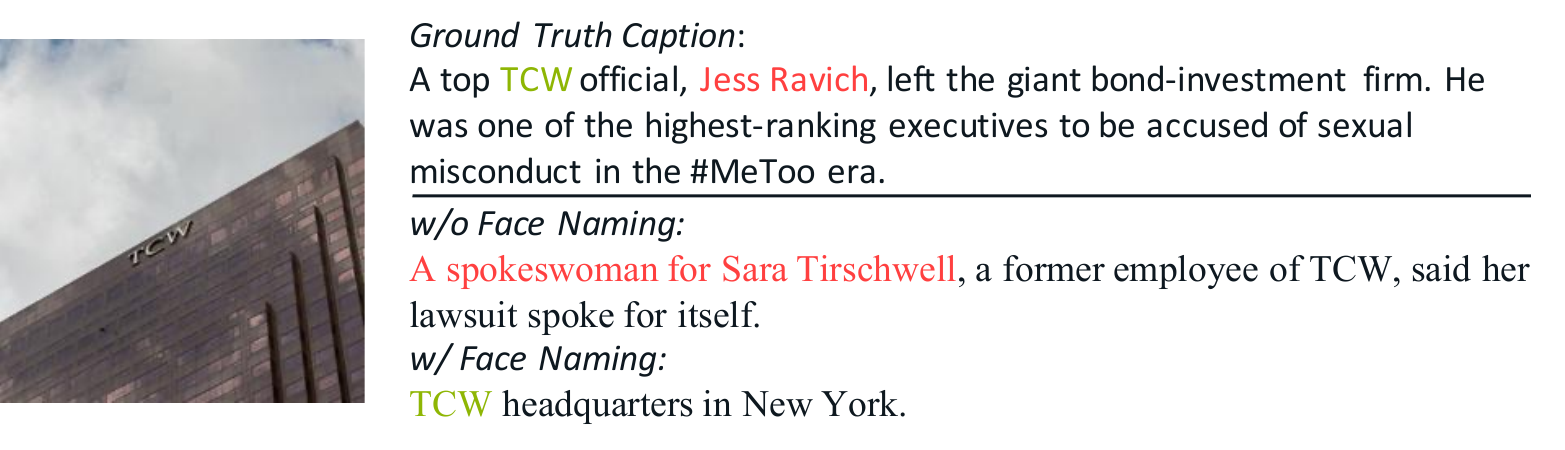}
\hdashrule[0.5ex][x]{\linewidth}{0.5pt}{1.5mm}
\includegraphics[width=\columnwidth]{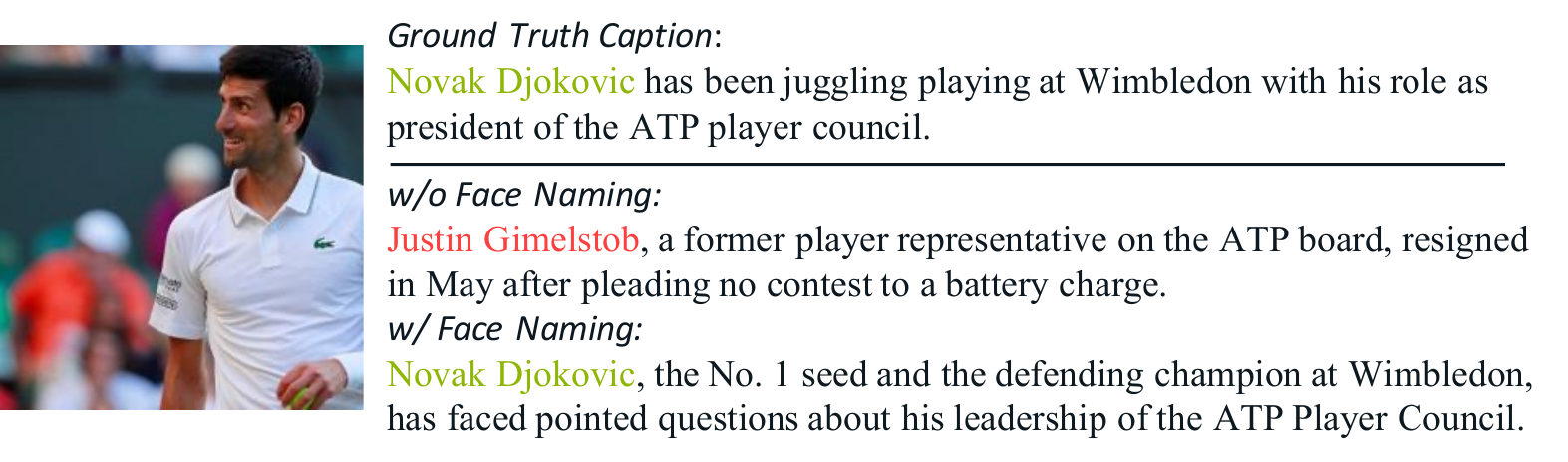}
\hdashrule[0.5ex][x]{\linewidth}{0.5pt}{1.5mm}
\includegraphics[width=\columnwidth]{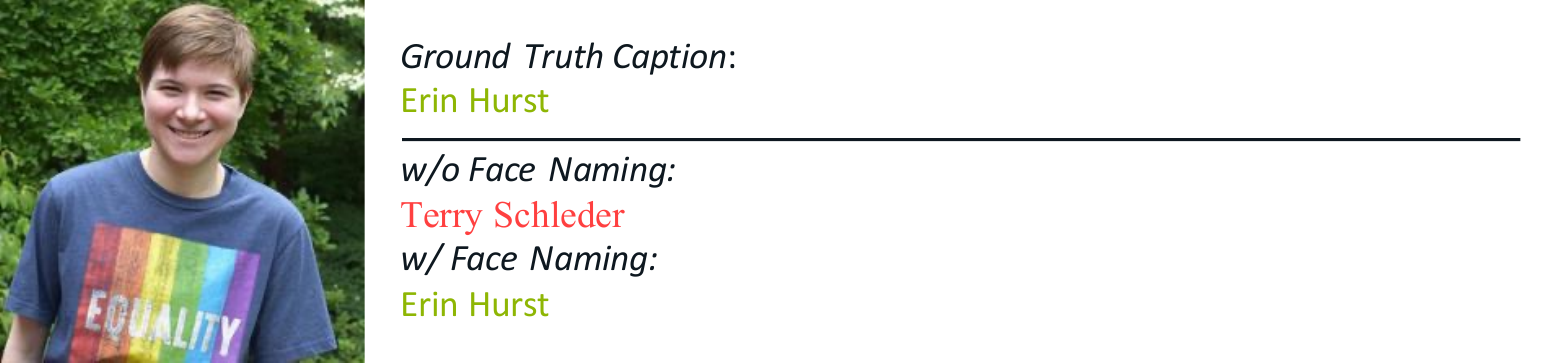}
\hdashrule[0.5ex][x]{\linewidth}{0.5pt}{1.5mm}
\caption{Qualitative comparison w/ or w/o Face Naming. For the correctly grounded person, we mark the names in \textcolor{applegreen}{green}; for the wrongly grounded person or person not appearing in the image, we mark the names in \textcolor{coralred}{red}. Here we use the models with $\text{BART}_{\text{base}}$ as backbone LM for comparison.}
\label{fig:gen_examples_facenaming}
\end{figure}

Then we qualitatively examine the effectiveness of the face naming module. We present multiple examples of generated captions using $\text{Ours}_{\text{base}}$ with or without the face naming module in Figure~\ref{fig:gen_examples_facenaming}. Without the face naming module, model tends to make mistakes by not identifying certain person (Figure~\ref{fig:gen_examples_facenaming}, first example) or grounding the person to wrong names (Figure~\ref{fig:gen_examples_facenaming} third and fourth examples). Interestingly, as shown in the second example of Figure~\ref{fig:gen_examples_facenaming}, name "Jess Ravich" appears in the ground truth caption, which corresponds with an image containing only a building. With the addition of the face naming module, the generated caption does not contain "Jess Ravich". Instead, it remains faithful by linking the "TCW" building to context in the article ("New York"). 
In contrast, without the face naming module, model generates caption with irrelevant names.
These qualitative examples prove the effectiveness of our face naming module.

\subsection{Qualitative Examples with The Addition of CoLaM}

\begin{figure}[!htbp]
\centering
\hdashrule[0.5ex][x]{\linewidth}{0.5pt}{1.5mm}
\includegraphics[width=\columnwidth]{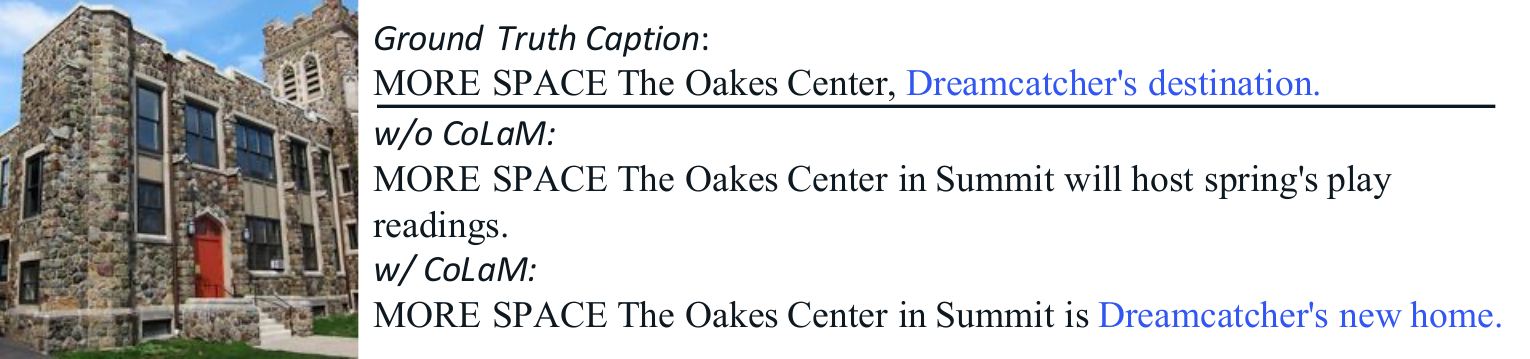}
\hdashrule[0.5ex][x]{\linewidth}{0.5pt}{1.5mm}
\includegraphics[width=\columnwidth]{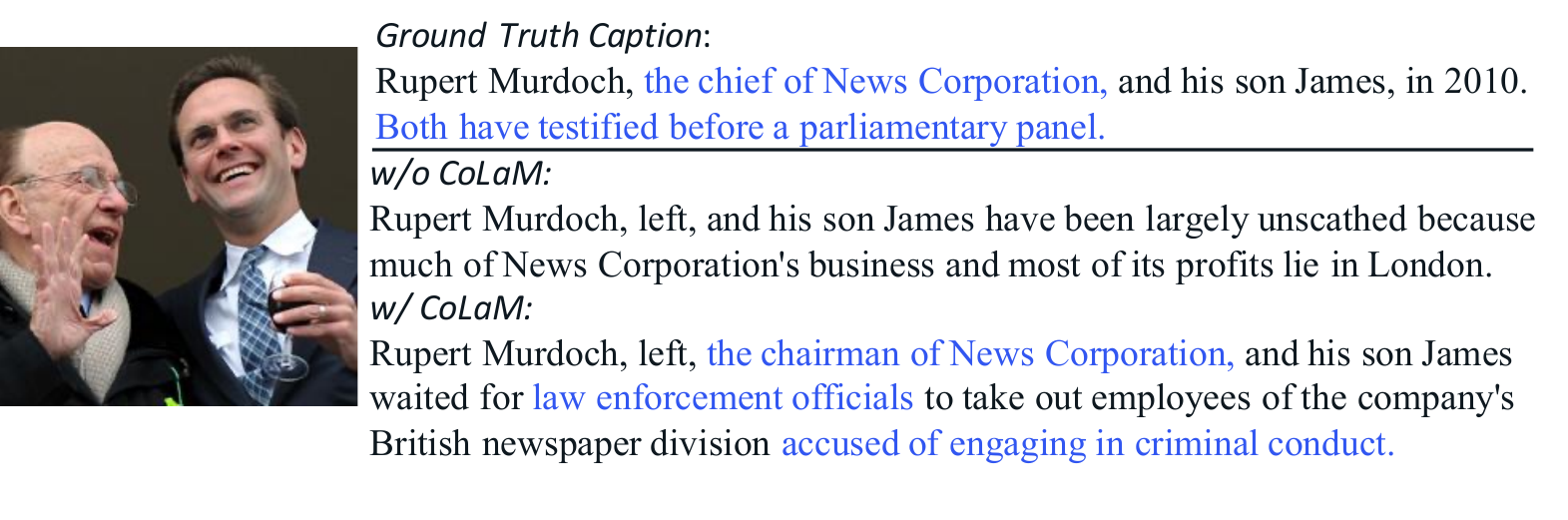}
\hdashrule[0.5ex][x]{\linewidth}{0.5pt}{1.5mm}
\includegraphics[width=\columnwidth]{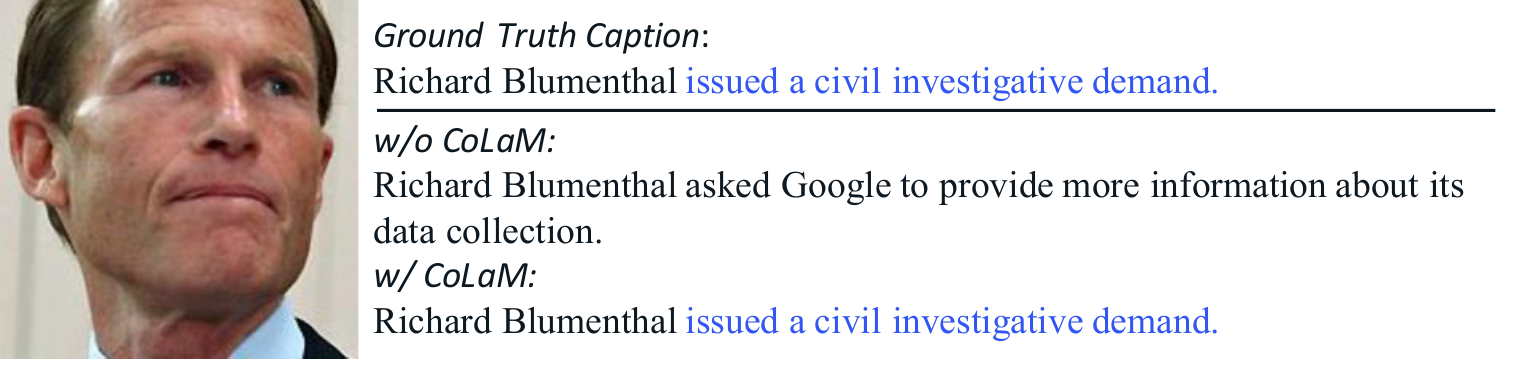}
\hdashrule[0.5ex][x]{\linewidth}{0.5pt}{1.5mm}
\includegraphics[width=\columnwidth]{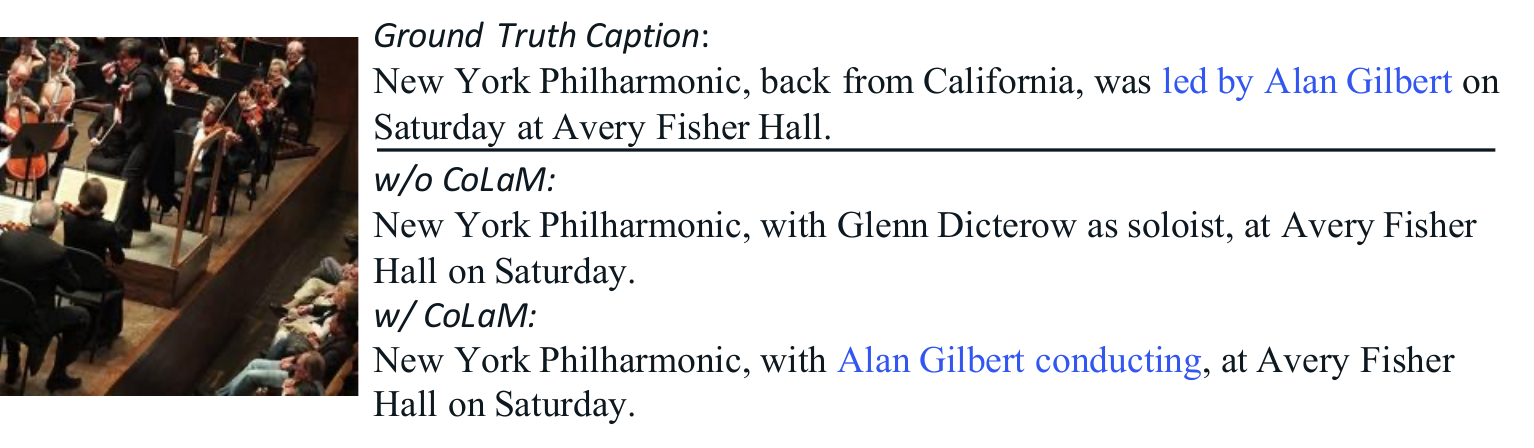}
\hdashrule[0.5ex][x]{\linewidth}{0.5pt}{1.5mm}
\caption{Qualitative comparison w/ or w/o CoLaM. We mark \textcolor{blue_ppt}{the context from the news articles that is captured by our method in blue}. Here we use the models with $\text{BART}_{\text{base}}$ as backbone LM for comparison.}
\label{fig:gen_examples_colam}
\end{figure}

We present multiple examples of captions generated using $\text{Ours}_{\text{base}}$ with or without CoLaM in Figure~\ref{fig:gen_examples_colam}.
With the addition of CoLaM, our model is able to capture the correct context from the article, which leads to improved captioning performance.

\begin{table*}
\centering
\resizebox{0.9\linewidth}{!}{%
\begin{tabular}{ m{0.15\linewidth}  m{0.27\linewidth}  m{0.5\linewidth} }
     \toprule
      Image & Article & Caption \\ 
      \hline
     \includegraphics[width=\linewidth, height=25mm]{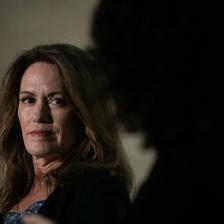}
      & \footnotesize
      Exposing the Bias Embedded in Tech...
      ...said \textcolor{applegreen}{Peggy Johnson, executive vice president of business development at Microsoft}... Ms. Johnson told an audience at \textcolor{applegreen}{the New Rules Summit}, ...
      
      \textcolor{coralred}{Meredith Whittaker, a founder and a director of the AI Now Institute at New York University}...said Ms. Whittaker, who also works at Google. \textcolor{coralred}{“It’s almost never white men... historical marginalization.”}...
      
      \textit{(Johnson occurs 4 times, Whittaker occurs 7 times in total.)}
      & 
      \begin{enumerate}[wide=0em, label=\Roman*, itemsep=0pt, parsep=0pt, font=\footnotesize\bfseries]
      \item \footnotesize \textbf{GT:} Peggy Johnson of Microsoft said biases needed to be exposed to be addressed. "The way to turn anything around is to shine a light on it."
      
      \item \footnotesize \textbf{(1) $\text{BART}_{\text{base}}$:} "It's almost never white men who are discriminated against by these systems," said \textcolor{coralred}{Meredith Whittaker}, \textcolor{coralred}{a founder and a director of the AI Now Institute at New York University}.
      
      \item \footnotesize \textbf{(2): (1) + visual features:} "It's almost never white men who are discriminated against by these systems," said \textcolor{coralred}{Meredith Whittaker}, \textcolor{coralred}{a founder and a director of the AI Now Institute at New York University}.
      
      \item \footnotesize \textbf{(3): (2) + face naming:} \textcolor{applegreen}{Peggy Johnson, executive vice president of business development at Microsoft}, spoke at \textcolor{applegreen}{the New Rules Summit}.
      
      \item \footnotesize \textbf{(4): (3) + retrieval:} \textcolor{applegreen}{Peggy Johnson, executive vice president of business development at Microsoft.}
      \end{enumerate}
      \\ 
      \midrule
      \includegraphics[width=\linewidth, height=25mm]{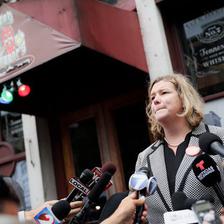}
      & \footnotesize President Plans Visits to Places... \textcolor{applegreen}{Dayton, Ohio,} and El Paso on Wednesday...
      \textcolor{applegreen}{The Democratic mayor of Dayton, Nan Whaley}, said on Tuesday that she had been  \textcolor{applegreen}{“disappointed” with Mr. Trump’s remarks the day before about the two massacres}, which left a combined 31 people dead...
      \textcolor{coralred}{Representative Veronica Escobar}, and her predecessor,...
      But like Ms. Whaley in Dayton, \textcolor{coralred}{El Paso’s mayor, Dee Margo}, a Republican ...
      & 
      \begin{enumerate}[wide=0em, label=\Roman*, itemsep=0pt, parsep=0pt, font=\footnotesize\bfseries]
      \item \footnotesize \textbf{GT:} Mayor Nan Whaley of Dayton said that she was "disappointed" with Mr. Trump's remarks about the two massacres but would welcome him to the city.
      
      \item \footnotesize \textbf{(1) $\text{BART}_{\text{base}}$:} \textcolor{applegreen}{President Trump} spoke at the \textcolor{coralred}{National Rifle Association convention} in \textcolor{coralred}{Louisville, Ky.}, on \textcolor{coralred}{Monday}.
      
      \item \footnotesize \textbf{(2): (1) + visual features:} \textcolor{coralred}{Representative Veronica Escobar}, \textcolor{coralred}{Democrat of El Paso}, has urged \textcolor{applegreen}{President Trump} not to visit the city. "He should not come here while we're in mourning," she said.
      
      \item \footnotesize \textbf{(3): (2) + face naming:} \textcolor{applegreen}{Nan Whaley, the mayor of Dayton}, Ohio, said on \textcolor{applegreen}{Tuesday} that she was \textcolor{applegreen}{"disappointed" with President Trump's remarks the day before about the two massacres.}
      
      \item \footnotesize \textbf{(4): (3) + retrieval:} \textcolor{applegreen}{Mayor Nan Whaley of Dayton}, Ohio, said on \textcolor{applegreen}{Tuesday} that she had been \textcolor{applegreen}{"disappointed" with President Trump's remarks the day before about the two massacres.}
      \end{enumerate}
      \\ 
      \bottomrule
\end{tabular}%
}
\caption{Generated captions with the models discussed in the ablation. 
We mark the correct/wrong content in \textcolor{applegreen}{green}/\textcolor{coralred}{red}. Here we adopt model w/o CoLaM for illustration. Qualitative examples for model w/ CoLaM can be found in Figure~\ref{fig:gen_examples_colam}.}
\label{tab:qualitative}
\end{table*}

\end{document}